\documentclass[twoside]{article}

\usepackage[accepted]{aistats2025}
\usepackage{microtype}
\usepackage{graphicx}
\usepackage{subfigure}
\usepackage{booktabs} 

\usepackage{hyperref}

\usepackage{enumitem}

\usepackage{amsmath,amssymb,mathtools,stmaryrd,amsthm,amsfonts}

\usepackage[capitalize,noabbrev]{cleveref}

\theoremstyle{plain}
\newtheorem{theorem}{Theorem}[section]
\newtheorem{proposition}[theorem]{Proposition}

\newtheorem{definition}[theorem]{Definition}

\newtheorem{remark}[theorem]{Remark}

\usepackage[textsize=tiny]{todonotes}
\usepackage{hyperref, xcolor}
\definecolor{linkcolor}{RGB}{83,83,182}
\hypersetup{
    colorlinks=true,
    citecolor=linkcolor,
    linkcolor=linkcolor
}
\usepackage{url}            
\usepackage{nicefrac}       
\usepackage{xcolor,pifont}        
\usepackage{subfigure}
\usepackage{bbm}
\usepackage{array,adjustbox}
\usepackage{natbib}
\usepackage{multirow}
\usepackage{multicol}
\usepackage{color}
\usepackage{caption}
\usepackage{changes}
\usepackage{floatrow}
\usepackage{algorithm,algorithmic}
\DeclareCaptionFormat{myformat}{#3}
\newcommand{\bm}[1]{\mathbf{#1}}

\begin{document}

%

%

\twocolumn[

\aistatstitle{Signature Isolation Forest}

\aistatsauthor{ Marta Campi$^{1}$  \And Guillaume Staerman$^{2}$ \And  Gareth W. Peters$^{3}$ \And  Tomoko Matsui$^{4}$ }

\aistatsaddress{ \footnotesize $^{1}$Institut Pasteur, Université Paris Cité, Inserm, Institut de l’Audition, IHU reConnect, Paris, France\\  $^{2}$INRIA, CEA, Univ. Paris-Saclay, France  \\ $^{3}$Department of Statistics \& Applied Probability, University of California Santa Barbara, USA \\ $^{4}$Institute of Statistical Mathematics, Tokyo, Japan } ]

\begin{abstract}
Functional Isolation Forest (FIF) is a recent state-of-the-art Anomaly Detection (AD) algorithm designed for functional data. It relies on a tree partition procedure where an abnormality score is computed by projecting each curve observation on a drawn dictionary through a linear inner product. Such linear inner product and the dictionary are a priori choices that highly influence the algorithm's performances and might lead to unreliable results, particularly with complex datasets. This work addresses these challenges by introducing \textit{Signature Isolation Forest}, a novel AD algorithm class leveraging the rough path theory's signature transform. Our objective is to remove the constraints imposed by FIF through the proposition of two algorithms which specifically target the linearity of the FIF inner product and the choice of the dictionary. We provide several numerical experiments, including a real-world applications benchmark showing the relevance of our methods.
\end{abstract}

\section{Introduction}
The development of sophisticated anomaly detection (AD) methods plays a central role in many areas of statistical machine learning, see \cite{Chandola}. The area of AD is growing increasingly complex, especially in large data applications such as in network science or functional data analysis. 

In the classical AD multivariate scenario, observations can be represented as points in $\mathbb{R}^d$ and various methodologies have been introduced in the literature. A model-based approach is favored when there is knowledge about the data-generating process \citep{rousseeuw1999fast}. Conversely, in numerous real-world applications where the underlying data system is unknown, nonparametric approaches are preferred \citep{polonik1997minimum,scholkopf2001estimating}. 

The area of AD is only just beginning to explore the domain of functional data. Functional data is becoming an important discipline in topological data analysis and data science, see \cite{ramsay2005fitting}; \cite{ferraty2006nonparametric} or \cite{wang2016functional} for more details. In essence, functional data involves treating observations as entire functions, curves, or paths. In some application domains, this approach can provide richer information than just a sequence of data observation vectors, yet it comes with challenges, especially when considering the development of AD methods.

In the use of AD within a functional data context, the anomaly detection task involves identifying which functions/curves differ significantly in the data to be considered anomalous in some sense. Following \cite{hubert2015multivariate}, three types of basic anomalies can be detected: \textit{shift}, \textit{shape},  \textit{amplitude}. These anomalies can then be \textit{isolated/transient} or \textit{persistent} depending on the frequency of their occurrences. Some are easier/more difficult to identify than others, e.g. an isolated anomaly in shape is much more complex than a persistent anomaly in amplitude.

At first sight, one may adopt a preliminary filtering technique. This leads to projecting the functional data onto a suitable finite-dimensional function subspace. It can be done by projecting onto a predefined basis such as Fourier or Wavelets or on a data-dependent Karhunen-Loeve basis by means of Functional Principal Component Analysis (FPCA; \citealp{dauxois1982asymptotic,shang2014survey}). The coefficients describing this subspace are then used as input into a specific anomaly detection algorithm designed for multivariate data. 

In this approach, the a priori choice of basis to undertake the projection can significantly influence the ability to detect particular types of anomalies in the functional data accurately. This makes these approaches sensitive to such choices. In other words, the choice of a given bases for projection can artificially amplify specific patterns or make others disappear entirely, depending on their properties. Therefore, it is a choice highly sensitive to the knowledge of the underlying studied process, which is a must for such an AD method to work.

If one seeks to avoid this projection methods that take functional data to a parameter vector space to undertake the AD method, one may alternatively seek AD methods working directly in the function space. This is the case of functional data depth, which takes as input a curve and a dataset of curves and returns a score indicating how deep the curve w.r.t. the dataset is. Many functional data depths, with different formulations, have been designed so far, such as, among others, the functional Tukey depth \citep{fraiman2001trimmed,claeskens2014multivariate}, the Functional Stahel-Donoho Outlyingness \citep{hubert2015multivariate}, the ACH depth \cite{staerman2020area}, the band depth \citep{lopez2009concept} or the half-region depth \citep{lopez2011half}; see \cite{nieto2016}; \cite{nagyfunctional} or the Chapter 3 of \cite{staerman2022functional} for a detailed review of functional depths.

One may also extend the classical anomaly detection algorithms designed for multivariate data such as the one-class SVM (OCSVM; \citealp{scholkopf2001estimating}), Local Outlier Factor (LOF; \citealp{Breunig}) or Isolation Forest (IF; \citealp{liu2008isolation}) to the functional setting, see \cite{rossi2006support}. Recently, \cite{staerman2019functional} introduced a novel algorithm, Functional Isolation Forest (FIF), which extends the popular IF to an infinite-dimensional case. It faces the challenge of adaptively (randomly) partitioning the functional space to separate trajectories in an iterative partition. Based on the computational advantages of random tree-based structures and their flexibility in choosing the split criterion related to functional properties of the data, it has been considered a powerful and promising approach for functional anomaly detection \citep{staerman2023}.

Although FIF offers a robust solution for functional data, three main challenges arise when it is applied in practice. Firstly, the selection of two critical parameters involved in the splitting criterion: the functional inner product and the dictionary of time-frequency atoms for data representation. The choice of these elements significantly influences the algorithm's performance, potentially compromising its flexibility. Secondly, the splitting criterion is formulated as a linear transformation of the dictionary function and a sample curve, potentially constrained in capturing more complex data, including non-stationary processes. Thirdly, when dealing with multivariate functions, FIF takes into account dimension dependency only linearly by using the sum of the inner product of each marginal as a multivariate inner product and may not capture complex interactions of the multivariate process. It is crucial to address these challenges to enhance the overall efficacy and adaptability of the FIF algorithm in practical applications.

In response to the above challenges, we propose a new class of algorithms, named \textsc{ (Kernel-) Signature Isolation Forest}, by leveraging the signature technique derived from rough path theory \citep{lyons2007differential,friz2010multidimensional}. In rough path theory, functions are usually referred to as paths to characterize their geometrical properties. The signature transform, or signature of a path, summarizes the temporal (or sequential) information of the paths. In practice, it captures the sequencing of events and the path's visitation order to locations. Yet, it entirely disregards the parameterization of the data, providing flexibility regarding the partial observation points of the underlying functions/paths. Consider a path in a multidimensional space of dimension $d$, i.e. $\mathbf{x} \in \mathcal{F}^d$, where the path consists of a sequence of data points. The signature of this path is a collection of iterated integrals of the path with respect to time. By including these iterated integrals up to a certain level, the signature provides a concise and informative representation of the path, allowing for extracting meaningful features. This makes it particularly useful in analyzing sequential data, such as time series. The idea behind such a transform is to produce a feature summary of the data system, which captures the main events of the data and the order in which they happen, without recording when they occur precisely \citep{lyons2022signature}. Performances of signature-based approaches appear to be promising, as they achieve state-of-the-art in some applications such as handwriting recognition \citep{wilson2018path,yang2016dropsample}, action recognition \citep{yang2016deepwriterid}, and medical time series prediction tasks \citep{morrill2019signature}. Surprisingly, it has been overlooked in the Machine Learning problems and has never been used in functional anomaly detection until we make this novel connection.

Our idea is to perform anomaly detection by isolating multivariate functions corresponding to sets of paths subject to the signature transform. Hence, the tree partitioning of FIF will now take place on the feature space of the signature transform. In such a way, the functional data changes are captured by embedding the studied data set into a path whose changes are then summarized by the signature method. We propose two different algorithms named \textsc{Kernel Signature Isolation Forest} (K-SIF) and \textsc{Signature Isolation Forest} (SIF). The first algorithm extends the FIF procedure to nonlinear transformations by using the signature kernel instead of the inner product of FIF. Relying on the attractive properties of the signature, see \cite{fermanian2021embedding,lyons2022signature} or Section \ref{sub:SIG}, our algorithms take into account higher moments of the functions (derivative, second-order derivative, etc.) hence accommodating changes (or oscillations) of different nature. The second algorithm is a more straightforward procedure relying on what is usually referred to as ‘coordinate signature', defined in Section \ref{sec:bck}, and being free of any sensitive parameters such as the dictionary (for FIF and K-SIF) and the inner product (for FIF). It removes the performance variability of the parameters used in FIF.

\begin{itemize}[noitemsep, topsep=0pt]\setlength\itemsep{1em}
    \item We introduce two functional anomaly detection algorithms (K-SIF and SIF) based on the isolation forest structure and the signature approach from rough path theory.
    
    \item The two algorithms provide improvements in the functional anomaly detection community by extending Functional Isolation Forest in two directions: the first one considers non-linear properties of the underlying data, hence providing a more suitable procedure for a more challenging dataset; the second one is an entirely data-driven technique free from any a priori choice due to a dictionary selection constraining the functional AD procedure to identify only specific types of patterns. 
    
    \item After studying the behavior of our class of algorithm regarding their parameters, we highlight the benefits of using (K-)SIF over FIF. On a competitive benchmark, we show that K-SIF, which extends the FIF procedure through the signature kernel, consistently showed the results of FIF on real-world datasets. Furthermore, we show that SIF achieves state-of-the-art performances while being more consistent than FIF.
\end{itemize}


\section{Background \& Preliminaries}\label{sec:bck}
Consider the functional random variable (r.v.) $\mathbf{X}$, taking values in the functional space $\mathcal{F}^d(I)$ of real valued functions (e.g. the Kolmogorov space $\mathcal{C}(I)$ or the Hilbert space $L_2(I)$) where $\textit{I} \subset \mathbb{R}$ represents an interval, such that
\begin{equation*}\label{eq:x_functional}
 \begin{split}
 \mathbf{X}: \Omega &\longrightarrow   \mathcal{F}^d(\textit{I}) \\
 \omega &\longmapsto \mathbf{X}(\omega) = (X_t (\omega))_{t \in \textit{I}}.
 \end{split}   
\end{equation*}
Note that throughout the rest of the paper, we will denote $\mathcal{F}^d(I) = \mathcal{F}^d$ for simplicity. Without loss of generality, we restrict our functions to be defined on $[0,1]$. In this paper, we assume $\mathcal{F}^d$ to be the space of bounded variation functions. A function $\mathbf{x} \in \mathcal{F}^d$ is said to be of bounded variation if for any discretization subset, the sum of differences is finite, i.e. $\underset{D}{\sup} \sum_{t_i \in D} || x_{t_i} - x_{t_{i-1}} || < \infty $ where $||\cdot ||$ is the Euclidean norm of $\mathbb{R}^d$ and the set discretization set $D$ is defined as
{\small 
\begin{equation*}
    D=\{ (t_0, \ldots, t_m) \;  | \;  0=t_0 < t_1 < \ldots < t_{k-1} < t_m = 1 \}.
\end{equation*}}
In practice, only a finite-dimensional marginal, $(X_{t_1}, X_{t_2}, \dots, X_{t_p})$  with $t_1 < \dots < t_p$ and $(t_1, \dots, t_p) \in [0,1]^p$, can be observed. To consider a function, rather than a set of discrete values, approximation or interpolation procedures are usually used and combined with a preliminary smoothing step when the observations are noisy. Denote by  $\mathcal{X}$ a reconstructed dataset of interpolated curves/functions $\mathcal{X}=\{\mathbf{x}_1, \ldots, \mathbf{x}_n \}$ from the set of observations $\{x_{i, t_1}, x_{i, t_2}, \dots, x_{i, t_p} \}_{i=1}^{n}$.
The problem of functional anomaly detection can be formulated as learning a score function $s: \mathcal{F}^d \rightarrow \mathbb{R}$ from $\mathcal{X}$ reflecting the degree of abnormality of any element of an infinite dimensional space $\mathcal{F}^d$ w.r.t. $\mathcal{X}$. 

\subsection{Functional Isolation Forest}\label{sub:FIF}
Consider $\mathcal{H}$ the functional Hilbert space equipped with a inner product $\langle ., .\rangle_{\mathcal{H}}$ such that any ${\bf x} \in \mathcal{H}$ is a real function defined on $[0,1]$.
A Functional Isolation Forest is created through an assembly of functional isolation trees (F-itrees). Each F-itree is constructed via a series of random splits from a subsample (of size $m$) of $\mathcal{X}_n$. The abnormality score for an observation $\mathbf{x}$ is then determined as a monotonically decreasing transformation of $\mathbf{x}$'s average depth across the trees. The core concept lies in the randomness of the splits, where an observation markedly different from others is more likely to be isolated from $\mathcal{X}_n$, appearing at shallower levels in the F-itrees. The F-itrees are built based on a predetermined dictionary $\mathcal{D}\subset \mathcal{H}$, encompassing both deterministic and/or stochastic functions capturing pertinent data properties, which may also be a subset of $\mathcal{X}_n$. Before each random univariate split, all node observations are projected onto a line defined by a randomly selected element from the dictionary $\mathcal{D}$. The selection of a suitable dictionary plays a pivotal role in shaping the FIF score construction. The projection criterion at each node of each F-itree is defined as:
\begin{equation*}
    \langle  \mathbf{x}, \mathbf{d} \rangle_{\mathcal{H}} = \alpha \times \dfrac{\langle  \mathbf{x}, \mathbf{d} \rangle_{_{L_2}}}{||\mathbf{x}|| ||\mathbf{d}||} + (1-\alpha) \times \dfrac{\langle  \mathbf{x'}, \mathbf{d'} \rangle_{_{L_2}}}{||\mathbf{x}'|| ||\mathbf{d}'||},
\end{equation*}
where $\mathbf{x}', \mathbf{d}'$ are derivative and $||\mathbf{x}||$ is the $L_2$ norm of $\mathbf{x}\in L_2([0, 1])$. When multivariate functional data are considered, the inner product employed is the sum of the marginal inner product accross the dimension i.e. $\langle \mathbf{x}, \mathbf{d} \rangle_{\mathcal{H}^d} = \sum_{i=1}^{d} \langle  \mathbf{x}^{(i)}, \mathbf{d}^{(i)} \rangle_{\mathcal{H}}$. The tree structure is equivalent to our algorithms and will be further detail in Section~\ref{sec:SIF}. For more detail about FIF specification, we refer the reader to \cite{staerman2019functional}.

\subsection{The Signature Method}\label{sub:SIG}
The signature of a path is a sequence of iterated integrals that captures important information about the path's geometric and topological features \citep{lyons2007differential,fermanian2021embedding}.
\begin{definition} Let $\mathbf{X}\in \mathcal{F}^d$ be a r.v. of bounded variation. For any set of coordinates $\{i_1,\ldots, i_k\} \subset \{1,\ldots, d \}^k, k\in \mathbb{N}_*$, and $[s, t]\subset [0, 1]$, the associated coordinate signature is defined by
\begin{align*}
    S_{(i_1, \ldots, i_k)}(\mathbf{X})_{[s, t]} &= \underset{s\leq u_1 < \ldots < u_k \leq t}{\idotsint}\mathrm{d}X_{u_1}^{i_1}\ldots \mathrm{d}X_{u_k}^{i_k}. 
\end{align*}
Furthermore, the signature of $\mathbf{X}$ is defined as the infinite collection of coordinate signature 
{\small
\begin{align*}
    S(\mathbf{X}) = \big (&1 , S_1(\mathbf{X}), \ldots, S_d(\mathbf{X}), \nonumber
    \\& S_{(1, 1)}(\mathbf{X}),S_{(1, 2)}(\mathbf{X}), \ldots ,S_{(d, d)}(\mathbf{X}), \nonumber
    \\&S_{(1,1, 1)}(\mathbf{X}),S_{(1,2, 1)}(\mathbf{X}), \ldots, S_{(d,d, d)} (\mathbf{X}), \ldots  \big ).
\end{align*}}
\end{definition}

\noindent \textbf{Truncated  Signature.}
In practice, to be computable the truncated signature is used. Given an order of truncation $k \in \mathbb{N}_*$, it is defined as the signature vector of finite length of dimension $C = \left|\{1,\ldots, d \}^k\right|$ given by
{\small
\begin{align*}\label{eq:sig2}
    S^k(\mathbf{X}) = \big (&1 , S_1(\mathbf{X}), \ldots, S_d(\mathbf{X}),
     S_{(1, 1)}(\mathbf{X}),S_{(1, 2)}(\mathbf{X}), \ldots , \\&S_{(d, d)}(\mathbf{X}),  \ldots, S_{\underbrace{(d,\ldots, d)}_{k}} (\mathbf{X})\big ) \in \mathbb{R}^{C}.
\end{align*}}

\noindent \textbf{Kernel Truncated Signature.} The signature can be seen as a feature map that embeds a function or a path into the tensor algebra. The truncated signature kernel has been recently introduced and studied in \cite{kiraly2019kernels}. It is defined as the map $K^k: \mathcal{F}^d \times \mathcal{F}^d \rightarrow \mathbb{R}$ such that
\begin{equation}\label{eq:kern_sig}
K^k(\mathbf{X}, \mathbf{Y}) = \langle S^k( \mathbf{X}), S^k( \mathbf{Y}) \rangle.    
\end{equation}

One may refer to \cite{lee2023signature} for an account of the untruncated kernel signature. See also Section~1 in the Appendix for further details about the signature and its properties.


\section{Signature Isolation Forest Methods}\label{sec:SIF}
With Kernel Signature Isolation Forest (K-SIF), we aim to leverage the truncated kernel signature \citep{kiraly2019kernels} to overcome the linearity constraint imposed by the inner product in FIF. In contrast to FIF, which explores only one function characteristic at each node using a unique inner product with the function sampled in the dictionary, K-SIF captures significantly more information. This is achieved by computing several coefficient signatures, summarizing multiple data attributes at each node. 

The ability to explore non-linear directions in the feature space through the kernel signature makes the algorithm more efficient for an equivalent computational cost (refer to Figure~6 in the Appendix).

\begin{figure}[!h]
\includegraphics[trim=1cm 0 0 0,height=4.5cm]{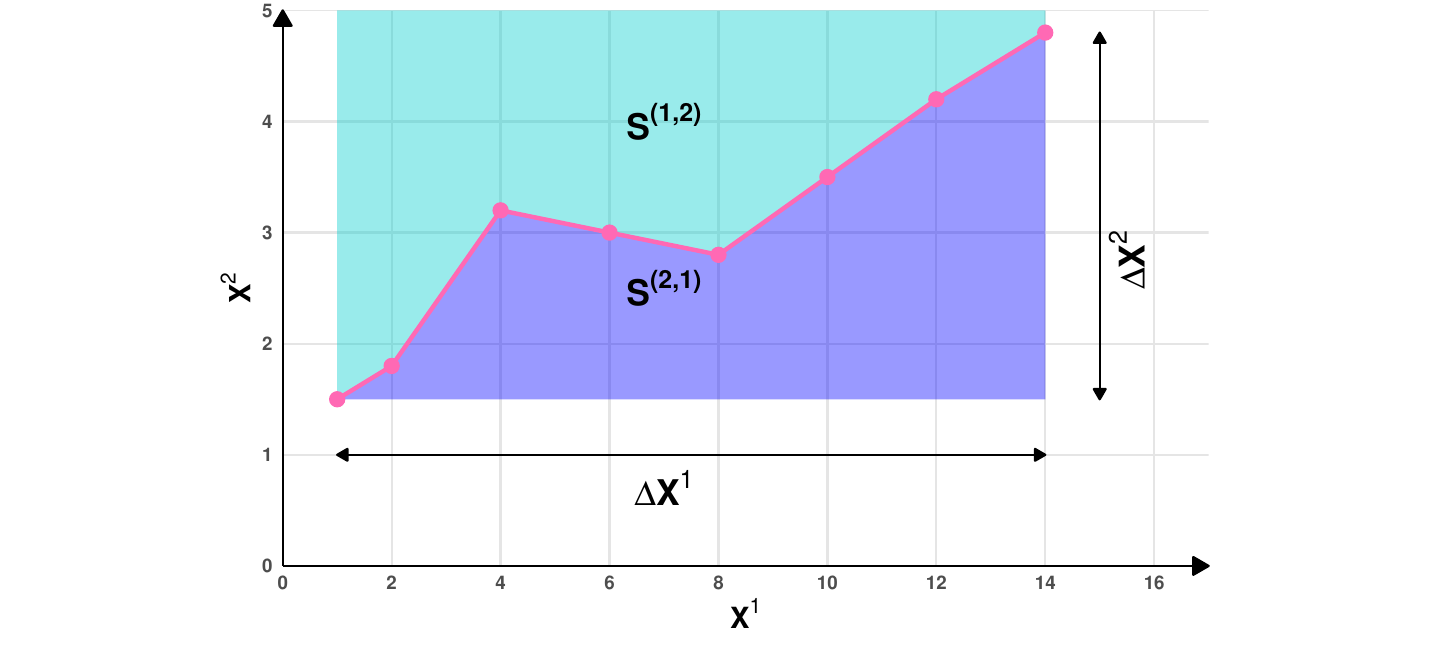}
\caption{Geometric visualization of depth-2 signature terms, where $S^{(1,2)}$ (cyan region) and $S^{(2,1)}$ (purple region) represent areas corresponding to coordinate signatures. The displacement term $\Delta X_1$ and $\Delta X_2$  along each axis capture the depth-1 terms of the transform.}
\label{fig:sig_def}
\end{figure}

With Signature Isolation Forest (SIF), the aim is to leverage the coordinate signature and remove any restrictions imposed by FIF and K-SIF. We replace the use of a dictionary $\mathcal{D}$ and rely on the coordinate signature only to detect anomalies. The intuition is to provide a more straightforward, entirely data-driven solution and remove any a priori required choice.

The construction procedures of (K-)SIF are now described. 

The objective of (K-)SIF  is to construct a collection of \textit{(kernel) signature isolation trees}, abbreviated by \textit{(k)si}-trees, obtained from $\mathcal{X} = \{ {\bf x }_1, \dots, {\bf x }_n \}$, a training sample composed of independent realizations of ${\bf X}$ taking values in the space $\mathcal{F}^d$.
If the dataset is too big, then a subsampling step can be performed to avoid masking or swamping effects \citep{Chandola}. This provides a subsample $\mathcal{X}_{\mathrm{sub}} = \{ {\bf x }_{i_{1}}, \dots, {\bf x }_{i_m} \}$ of size $m$ drawn randomly and uniformly to construct each tree. The global structure of a tree remains equivalent whether for a \textit{ksi}-tree or a \textit{si}-tree and is formally defined as follows.

\begin{definition}
A \textit{(k)si}-tree $\tau$ of depth $P \geq 1$ is a binary tree corresponding to a nested sequence of partitions of the functional space $\mathcal{F}^d$.   
\end{definition}

\noindent \textbf{Tree Construction.}  Each \textit{(k)si}-tree is obtained by recursively filtering $\mathcal{X}_{\mathrm{sub}}$,  by means of a signature-based criterion, in a top-down fashion according to the procedure here presented. Such a tree starts with a \textit{root node}, or initial node, corresponding to the entire feature space and given by $\mathcal{C}_{0,0} = \mathcal{F}^d$. Any other node encountered during the procedure will be denoted by the pair $(p,q)$ where $p$ is the index of the depth of the node (i.e. at which splitting node the algorithm has branched) with $0 \leq p < P$, while $q$ is linked to the subset $\mathcal{C}_{p,q} \subset \mathcal{F}^d$ with $0 \leq q \leq 2^p - 1$. A non-terminal node $(p,q)$ has two children corresponding to two disjoint subset $\mathcal{C}_{p+1,2q}$ and $\mathcal{C}_{p+1,2q+1}$ such that $\mathcal{C}_{p,q} = \mathcal{C}_{p+1,2q} \cup \mathcal{C}_{p+1,2q+1}$. A node $(p,q)$ is said to be terminal if it has no children.
 
\noindent \textbf{Splitting Criterion.} The difference between a \textit{ksi}-tree and a \textit{si}-tree lies in the splitting criterion computed at each node of both trees. Below, we detail the splitting procedures occurring at a particular internal node for both algorithms.

\noindent \textbf{Kernel Signature Isolation Forest.} Due to the truncated signature use, the depth of this truncation level $k$ must be a priori chosen. Whilst the SIF does not require a dictionary specification, the K-SIF approaches algorithm involves as parameters a dictionary $\mathcal{D} \subset \mathcal{F}^d$, which is chosen to be rich enough to represent properties of the data considered and a probability measure $\boldsymbol{\nu}$ on $\mathcal{D}$. Here we use the three standard dictionaries of FIF \citep{staerman2019functional}: ‘Brownian', a dictionary of standard Brownian motion paths (corresponding to the space of continuous functions with the Wiener measure $\boldsymbol{\nu}$), ‘Gaussian wavelets' representing a Mexican hat wavelet basis and ‘Cosine' that is a cosine basis. In the latter two dictionary cases the probability measure $\boldsymbol{\nu}$ is usually uniform over the class of dictionary functions, which is suitable as an uniformative measure a priori.

The dataset at node $(p,q)$ is denoted at $\mathcal{X}_{p,q}$. At iteration $q+2^p$ of the \textit{ksi}-tree construction, a split variable is selected by drawing a function $\mathbf{d}$  from $\mathcal{D} \subset \mathcal{F}^d$ according to a distribution $\boldsymbol{\nu}$. The data are then projected on such dictionary but through the signature kernel embedding, i.e. for a given ${\bf d} \in \mathcal{D}$ (where ${\bf d}$ represents a function sampled from $\mathcal{D}$ at each split) the projection of a sole function ${\bf x}$ considering the kernel signature is given according to the truncated signature kernel defined earlier in Eqn. \ref{eq:kern_sig} applied to $\bm{x}$ and dictionary item $\mathbf{d}$.This is performed $\forall {\bf x} \in \mathcal{X}_{p, q}$ and $k$ is a priori chosen. 

The following step is to choose uniformly and at random a split value $\gamma$ such that  
\begin{equation*}
\gamma \in \left[ \min_{ {\bf x} \in \mathcal{X}_{p,q} } \langle S^k ({\bf x}), S^k ({\bf d}) \rangle  , \; \; \max_{ {\bf x} \in \mathcal{X}_{p,q} } \langle S^k ({\bf x}), S^k ( {\bf d} ) \rangle   \right].
\end{equation*}
Then the algorithm can split $\mathcal{X}_{p,q}$ and generate a new node where the  the children subsets are then defined by $\mathcal{C}_{p+1,2q} = \mathcal{C}_{p,q} \cap C_{\text{K-SIF}}^L $ and $\mathcal{C}_{p+1,2q+1} = \mathcal{C}_{p,q} \cap C_{\text{K-SIF}}^R$ with $L$ and $R$ indicating left and right subset splits given as
\begin{equation*}\label{eq:criteria_KSIF}
\begin{split}
C_{\text{K-SIF}}^L = &\{ {\bf x} \in \mathcal{F}^d: \langle S^k ({\bf x}), S^k ({\bf d}) \rangle \leq \gamma \}\\
C_{\text{K-SIF}}^R = &\{ {\bf x} \in \mathcal{F}^d: \langle S^k ({\bf x}), S^k ({\bf d}) \rangle > \gamma \}.
\end{split}
\end{equation*}
The children datasets are given as 
\begin{equation*}
\begin{split}
\mathcal{X}_{p+1,2q} &= \mathcal{X}_{p,q} \cap \; \mathcal{C}_{p+1,2q} \\
\mathcal{X}_{p+1,2q+1} &= \mathcal{X}_{p,q} \cap \; \mathcal{C}_{p+1,2q+1}.
\end{split}
\end{equation*}
A \textit{ksi}-tree is built by iterating this procedure until all training data curves are isolated. 

\noindent \textbf{Signature Isolation Forest.}  
In contrast to \textit{ksi}-trees, the definition of a \textit{si}-tree will rely on using a different split criterion, expressed according to the following procedure. Once again, the depth of the truncation level $k$ has to be chosen a priori, but no dictionaries nor distribution has to be set. Here, at each internal node $(p,q)$, a coordinate $(i_1,\ldots, i_\ell)$ is chosen randomly and uniformly in the set 
$$\{ (i_1, \ldots
, i_\ell) \in \llbracket  1, d \rrbracket^\ell; \quad 1\leq \ell \leq k \}.$$

Note that the coordinates may be chosen according to a specific law with additional a priori knowledge of the data, allowing the coordinate signature to discriminate particular aspects of the functions. Thereafter, this coordinate signature is computed across $\mathcal{X}_{p,q}$ and a split value $\gamma$ is chosen uniformly such that 
\begin{equation*}
\gamma \in \left[ \min_{ {\bf x} \in \mathcal{X}_{p,q} } S_{(i_1,\ldots, i_\ell)}(\mathbf{x})  , \; \; \max_{ {\bf x} \in \mathcal{X}_{p,q} } S_{(i_1,\ldots, i_\ell)}(\mathbf{x})   \right].   
\end{equation*}
Then the algorithm can split $\mathcal{X}_{p,q}$ and generate a new node where the  the children subsets are then defined by $\mathcal{C}_{p+1,2q} = \mathcal{C}_{p,q} \cap C_{\text{SIF}}^L $ and $\mathcal{C}_{p+1,2q+1} = \mathcal{C}_{p,q} \cap C_{\text{SIF}}^R$ with $L$ and $R$ indicating left and right subset splits given as
\begin{equation*}\label{eq:criteria_SIF}
\begin{split}
C_{\text{SIF}}^L =& \{ {\bf x} \in \mathcal{F}^d:  S_{(i_1, \dots, i_\ell)} ({\bf x})  \leq \gamma \}\\
C_{\text{SIF}}^R =& \{ {\bf x} \in \mathcal{F}^d: S_{(i_1, \dots, i_\ell)} ({\bf x})  > \gamma \}.
\end{split}
\end{equation*}
The children datasets are given as 
\begin{equation*}\label{eq:child_data_KSIF}
\begin{split}
\mathcal{X}_{p+1,2q} &= \mathcal{X}_{p,q} \cap \; \mathcal{C}_{p+1,2q} \\
\mathcal{X}_{p+1,2q+1} &= \mathcal{X}_{p,q} \cap \; \mathcal{C}_{p+1,2q+1}.
\end{split}
\end{equation*}
A \textit{si}-tree is also built by iterating this procedure until all training data curves are isolated.  The two algorithms of K-SIF and SIF are summarized in Section 1 in the Appendix.

\noindent \textbf{Anomaly Score.} As the terminal nodes of a \textit{(k)si}-trees $\tau$ form a partition of the feature space, we can define the piecewise constant function $h_{\tau}: \mathcal{F}^d \rightarrow \mathbb{N}$ by: 
\begin{equation*}
h_{\tau}({\bf x}) = p \; \text{iff}  \;  {\bf x} \in \mathcal{C}_{p,q}  \;   \text{and}  \;  (p,q)  \;  \text{is a terminal node.}
\end{equation*}

This random path length offers an indication for the degree of abnormality: the more abnormal ${\bf x}$ is, the higher the probability that the quantity $h_{\tau}({\bf x})$ is small.
K-SIF (resp. SIF) build a collection $\tau_1, \ldots, \tau_N$ of $N$ ksi-trees (resp. si-tree). 
Given a $\mathbf{x} \in \mathcal{F}^d$, following \cite{liu2008isolation} and \cite{staerman2019functional}, we can define the monotone transformation of the averaged path length over the trees:
\begin{equation*}\label{eq:score_anomaly}
 s_n( {\bf x} ) = 2^{-\frac{1}{N c (m)} \sum_{l = 1}^{N} h_{\tau_l}({\bf x})},  
\end{equation*}
where $c (m)$ is the average path length of unsuccessful searches in a binary search tree and $m$ the size of the subsample linked to each tree.

\noindent \textbf{Parameters of K-SIF and SIF.} 
For both algorithms, key parameters typical of isolation forest-based methods, such as the number of trees $N$ or the subsample size $m$, must be pre-selected \citep{liu2008isolation}. The truncated level $k$ of the signature depth must also be chosen. In the case of K-SIF, similar to FIF, selecting a dictionary $\mathcal{D}$ and a distribution $\boldsymbol{\nu}$ is required. We opt to implement three standard dictionaries (also utilized in FIF): ‘Brownian', representing a traditional Brownian motion path; ‘Cosine', employing a cosine basis; and ‘Gaussian wavelets', utilizing a Gaussian wavelets basis.
In contrast, SIF does not require any of these sensitive parameters. We follow the approach outlined in \cite{morrill2019signature} to enhance the algorithms' performance by introducing a split window parameter $\omega$. This implies that, at each tree node, the truncated signature is computed on a randomly selected portion of the functions with a size of $\lfloor p / \omega \rfloor$.


\section{Numerical Experiments}\label{sec:num-exp}
This section presents a series of numerical experiments supporting the proposed class of AD methods. We organize the experiments into three categories. First, a parameter sensitivity analysis sheds light on the behavior of the parameters algorithms and provides insights for achieving optimal performances. Subsequently, we compare (K-)SIF and FIF, illustrating the validity of such methods and showcasing the power of the signature in the context of AD solutions. Finally, we perform a benchmark that compares several existing AD methods on various real datasets. This benchmark aims to demonstrate the results in real applications and assess the overall performance of these algorithms relative to other AD methods documented in the literature.

\subsection{Parameters Sensitivity Analysis}\label{sub:parameters}
We investigate the behavior of K-SIF and SIF with respect to their two main parameters: the depth of the signature $k$ and the number of split windows $\omega$. For the sake of place, the experiment on the depth is postponed in Section~3.1 in the Appendix.

\noindent \textbf{The Role of the Signature Split Window.} The number of split windows allows the extraction of information over specific intervals (randomly selected) of the underlying data. Thus, at each tree node, the focus will be on a particular portion of the data, which is the same across all the sample curves for comparison purposes. This approach ensures that the analysis is performed on comparable sections of the data, providing a systematic way to examine and compare different intervals or features across the sample curves. 

We explore the role of this parameter with two different datasets that reproduce two types of anomaly scenarios. The first considers isolated anomalies in a small interval, while the second contains persistent ones across all the function parametrization. In this way, we observe the behavior of K-SIF and SIF with respect to different types of anomalies.

The first dataset is constructed as follows. We simulate 100 constant functions. We then select at random 90\% of these curves and Gaussian noise on a sub-interval; for the remaining 10\% of the curves, we add Gaussian noise on another sub-interval, different from the first one. More precisely: 90\% of the curves, considered as normal, are generated according to    
    $\mathbf{x}(t)=b+ \varepsilon (t)\mathbb{I}(t\in [0.3,0.6]),$    
    with $\varepsilon (t) \sim \mathcal{N}(0, 1)$, $b \sim  \mathcal{U}([0, 100])$ and $\mathcal{U}$ representing the uniform distribution;
    10\% of the curves, considered as abnormal, are generated according to 
    $\mathbf{x}(t)=b+ \varepsilon (t)\mathbb{I}(t\in [0.7,0.8]),$
    where $\varepsilon(t) \sim \mathcal{N}(0, 1)$ and $b \sim  \mathcal{U}([0, 100])$.

The second dataset is constructed as follows. We generate 100 one-dimensional Brownian motion paths. We remark that a stochastic process $B_t$ follows a Brownian motion if it satisfies the following stochastic differential equation $d B_t = \mu dt + \sigma d X_t, $ where $X_t$ is a Wiener process or Brownian motion (with distribution $\mathcal{N}(0,1)$).

\begin{figure}[!ht]
    \centering
    \begin{tabular}{cc}
         \includegraphics[trim=0cm 0 0 0,height=4cm]{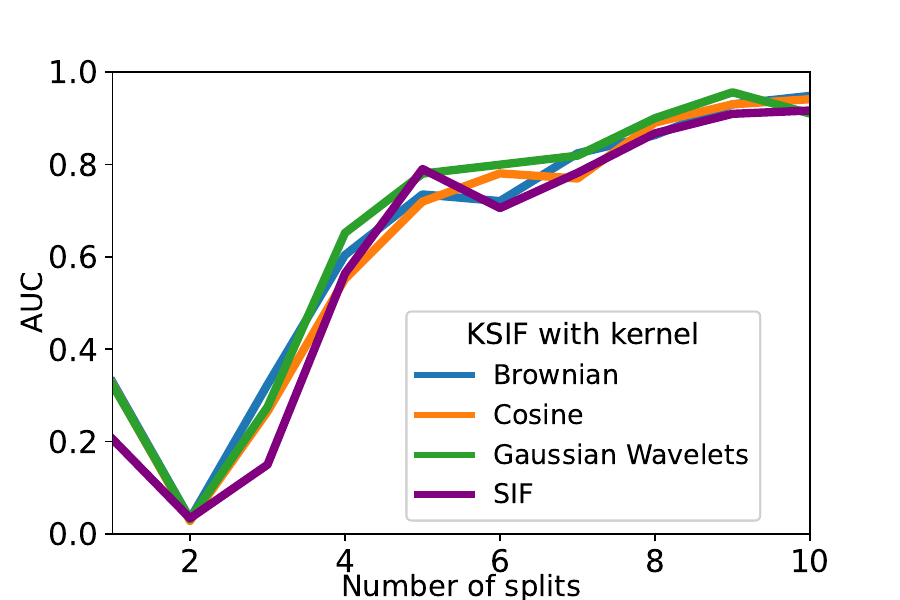} \\
         \includegraphics[trim=0cm 0 0 0,height=4cm]{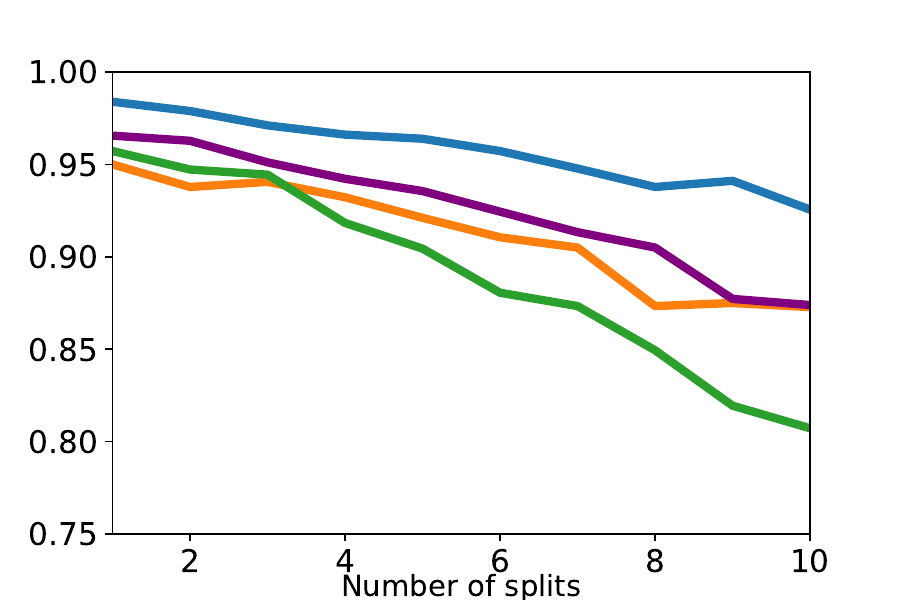}
    \end{tabular}
    \caption{AUC for the ROC curve w.r.t. the number of split window on the first (top) and the second (bottom) datasets for the three dictionaries.}
    \label{fig:split}
\end{figure}

We simulate at random 90\% of the paths with $\mu = 0$, $\sigma = 0.5$, and consider them as normal data. Then, the remaining 10\% are simulated with drift $\mu = 0.2$, standard deviation $\sigma = 0.4$, and considered abnormal data. We compute K-SIF with different numbers of split windows, varying from 1 to 10, with a truncation level set equal to 2 and $N=1,000$ the number of trees. The experiment is repeated $100$ times, and we report the averaged AUC under the ROC curves in Figure~\ref{fig:split} for both datasets and three pre-selected dictionaries.

For the first dataset, where anomalies manifest in a small portion of the functions, increasing the number of splits significantly enhances the algorithm's performance in detecting anomalies. The performance improvement shows a plateau after nine split windows. In the case of the second dataset with persistent anomalies, a higher number of split windows has a marginal impact on the algorithm's performance, maintaining satisfactory results. Therefore, without prior knowledge about the data, opting for a relatively high number of split windows, such as 10, would ensure robust performance in both scenarios. Additionally, a more significant number of split windows enables the computation of the signature on a smaller portion of the functions, leading to improved computational efficiency.

\subsection{(K-)SIF detects Swap Order Variation Changes}\label{subsec:advantages}

 The signature method considers the order of occurring events in functional data. To investigate this phenomenon with our proposed class of algorithms, we define a synthetic dataset of $100$ smooth functions given by
$\mathbf{x}(t)=30 t^q(1-t)^q, $
with $t\in [0, 1]$ and $q$ equispaced in $[1, 1.4]$. Then, we simulate the occurrences of events by adding Gaussian noise on different portions of the functions. We randomly select 90\% of them and add Gaussian values on a sub-interval, i.e., $\mathbf{x}(t)=30 t^q(1-t)^q + \varepsilon (t) \mathbb{I}(t\in [0.2, 0.4]),$
where  $\varepsilon(t) \sim \mathcal{N}(0, 0.8)$. We consider the 10\% remaining as abnormal by adding the same ‘events' on another sub-interval compared to the first one, i.e., $\mathbf{x}(t)=30 t^q(1-t)^q + \varepsilon (t) \mathbb{I}(t\in [0.6, 0.8]),$
where  $\varepsilon (t) \sim \mathcal{N}(0, 0.8)$. See Figure~5 in the Appendix for a dataset visualization. To summarize, we have constructed two identical events occurring at different parts of the functions, leading to \textit{isolating} anomalies.  As presented in the introduction, this class of anomalous data are amongst the most challenging to identify.

\begin{figure}[!ht]
\includegraphics[trim=0cm 0 0 0,height=4.5cm]{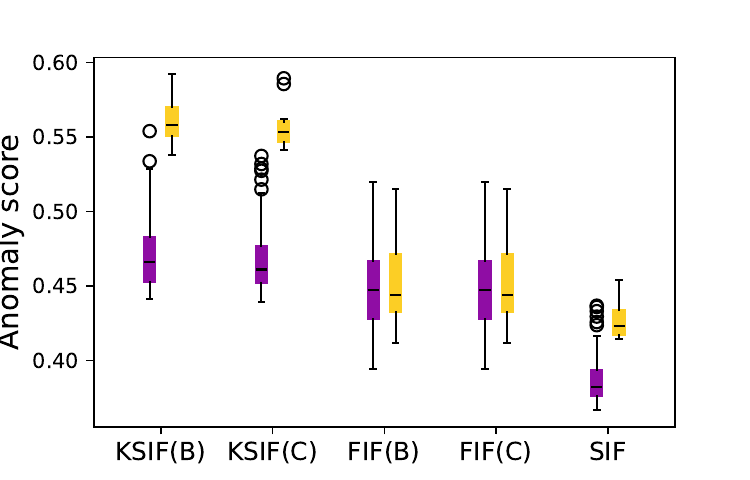}
\caption{Anomaly score for normal (purple) and abnormal  (yellow) data for SIF,  K-SIF and FIF with Brownian and Cosine dictionaries.}
\label{fig:swap}
\end{figure}

 We compute SIF, K-SIF and FIF with Brownian and Cosine dictionaries on these simulated datasets. As indicated in Section~\ref{sub:parameters}, for (K-)SIF, we choose $\omega = 10$, the number of split windows, and $k=2$, the depth of the signature. In Figure~\ref{fig:swap}, we report boxplots of the anomaly score returned by the algorithms for the normal data in purple and abnormal data in yellow. While this could appear a very simple task, in practice, it is highly challenging for an AD algorithm to differentiate such classes of curves. The introduction of the signature method should tackle precisely this type of scenario, since taking into account the order of the events.


{\renewcommand{\arraystretch}{1.5} 
{\setlength{\tabcolsep}{0.135cm} 
\begin{table*}[!ht]
    \caption{AUROC of different anomaly detection methods calculated on the test set. Bold numbers correspond to the best result (higher is better).}
    \vspace{-0.2cm}
    \centering
    {\scriptsize
    \begin{tabular}{|c||c|c|c|c||c|c|c|c|c|c|c|c|}
        \hline
        Methods: & SIF & K-SIF$_{\mathrm{GW}}$ & K-SIF$_{\mathrm{C}}$ & K-SIF$_{\mathrm{B}}$  & IF & OCSVM & fHD & fSDO & DeepSVDD & AnoGan & AutoEncoder & VAE \\
        \hline
        Chinatown & \textbf{1} & 0.90 & 0.99 & \textbf{1} & 0.81 & 0.6 & 0.73 & \textbf{1} & 0.5 & 0.72 & \textbf{1} & 0.99 \\
        \hline
        Coffee & 0.84 & \textbf{0.92} & 0.85 & 0.83 & 0.74 & 0.76 & 0.89 & 0.9 & 0.55 & 0.5 & \textbf{0.83} & \textbf{0.83} \\
        \hline
        ECGFiveDays & 0.93 & 0.90 & 0.92 & 0.90 & 0.85 & 0.93 & 0.84 & \textbf{0.96} & 0.67 & 0.79 & \textbf{0.96} & \textbf{0.96} \\
        \hline
        ECG200 & 0.85 & 0.82 & 0.85 & 0.83 & 0.85 & 0.81 & 0.83 & \textbf{0.89} & 0.5 & 0.72 & \textbf{0.87} & 0.86 \\
        \hline
        Handoutlines & \textbf{0.84} & 0.83 & 0.83 & 0.82 & \textbf{0.84} & 0.83 & 0.82 & 0.83 & 0.63 & 0.55 & \textbf{0.82} & \textbf{0.82} \\
        \hline
        SonyRobotAI1 & \textbf{0.99} & 0.96 & 0.95 & 0.95 & 0.96 & 0.96 & 0.88 & 0.96 & 0.52 & 0.91 & \textbf{0.95} & \textbf{0.95} \\
        \hline
        SonyRobotAI2 & \textbf{0.93} & 0.89 & 0.92 & \textbf{0.93} & 0.87 & 0.84 & 0.84 & 0.88 & 0.64 & 0.87 & \textbf{0.88} & \textbf{0.88} \\
        \hline
        StarLightCurves & \textbf{0.80} & 0.75 & 0.76 & 0.76 & 0.72 & 0.71 & 0.71 & 0.79 & 0.60 & 0.30 & 0.76 & \textbf{0.77} \\
        \hline
        TwoLeadECG & 0.92 & 0.92 & 0.92 & 0.92 & 0.78 & 0.54 & 0.63 & \textbf{1} & 0.46 & 0.68 & \textbf{1} & \textbf{1} \\
        \hline
        ECG5000 & 0.90 & 0.92 & \textbf{0.97} & 0.91 & 0.94 & 0.94 & \textbf{0.97} & 0.92 & 0.59 & 0.65 & \textbf{0.92} & \textbf{0.92} \\
        \hline
    \end{tabular}}
    \label{tab:bench}
\end{table*}}}
{\renewcommand{\arraystretch}{1.5} 
{\setlength{\tabcolsep}{0.25cm}
\begin{table*}[!htt]
\centering
{\scriptsize
\begin{tabular}{|l|c|c|c|c|c|c|c|c|c|c|c|}
\hline
Methods: & SIF & KSIF & FIF & IF & OCSVM & fHD & fSDO & DeepSVDD & AnoGan & AutoEncoder & VAE \\
\hline
Chinatown & 0.3 & 0.5 & 1.3 & 0.3 & 0.3 & 0.1 & 0.1 & 18 & 78 & 91 & 78  \\
\hline
Coffee & 0.8 & 1 & 1.6 & 0.3 & 0.3 & 0.1 & 0.1 & 16 & 88 &91 & 90 \\
\hline
ECGFiveDays & 0.6 & 0.6 & 1.2 & 0.5 & 0.6 & 0.3 & 0.3 & 21 & 84 & 87 & 89 \\
\hline
ECG200 & 3 & 5 & 9 & 1 & 1 & 0.7 & 0.7 & 18 & 151 & 91 & 90 \\
\hline
Handoutlines & 30 & 50 & 90 & 7 & 7 & 2 & 2 & 27 &681 & 593 & 365 \\
\hline
SonyRobotAI1 & 0.8 & 0.8 & 1.6 & 0.5 & 0.6 & 0.3 & 0.3 & 18 &84 & 95 & 84  \\
\hline
SonyRobotAI2 & 0.9 & 0.8 & 1.6 & 0.5 & 0.6 & 0.3 & 0.3 & 19 & 86& 92 &82   \\
\hline
StarLightCurves & 13 & 20 & 62 & 3 & 5 & 1 & 1 & 22 & 500& 202& 175 \\
\hline
TwoLeadECG & 0.6 & 0.6 & 1 & 0.3 & 0.3 & 0.2 & 0.2 & 9 & 75 &66 & 58  \\
\hline
ECG5000 & 5.4 & 4 & 15 & 1 & 1 & 0.6 & 0.6  & 9 & 270 & 102 & 91\\
\hline
\end{tabular}}
\caption{Computational time in seconds of different anomaly detection methods calculated on the test set.}
\label{tab:computational_time}
\end{table*}}}

Using both dictionaries, FIF fails to detect this anomaly as it is not designed to handle these type of phenomena. In contrast, K-SIF and SIF produce significantly distinguished scores between normal and abnormal data, efficiently classifying the second data class as anomaly. 

\subsection{Real-data Anomaly Detection Benchmark}\label{subsec:anom}
To evaluate the effectiveness of the proposed (K-)SIF algorithms and provide a comparison with FIF, we perform a comparative analysis using ten anomaly detection datasets constructed in \cite{staerman2019functional} and sourced from the UCR repository \citep{UCRArchive}. In contrast to \cite{staerman2019functional}, we do not use a training/test part since the labels are not used for the training and train and evaluate models on the training data only. We evaluate the algorithms' performance by quantifying the Area under the ROC curves. For completeness, the FPR at 95\% TPR and the Area under the PR curves are given in Tables 2 and 3 in the Appendix.

K-SIF  is examined in the context of three introduced dictionaries: Brownian, Cosine and Gaussian wavelets, as done for FIF. The parameters are configured with $N = 100$, $m = \text{min}(256,n)$, the height limit set to $\lceil \log_2 (m) \rceil$ for both FIF and K-SIF/SIF. The number of split windows is fixed with $\omega=10$, and the depth is fixed with $k=3$. We compare the K-SIF/SIF methods against two widely used multivariate anomaly detection techniques and two functional depths with default settings. The multivariate methods, namely, Isolation Forest (IF; \citealp{liu2008isolation}) and one-class support vector machine (OCSVM; \citealp{SPSSW01}), are applied following dimension reduction via Functional PCA. We retain 20 principal components with the largest eigenvalues using the Haar basis. The depths considered include the random projection halfspace depth \citep{cuevas} and the functional Stahel-Donoho outlyingness \citep{hubert2015multivariate}. 

\begin{figure}[!h]
    \centering
    \begin{tabular}{cc}
         \includegraphics[trim=0cm 0 0 0,height=5cm]{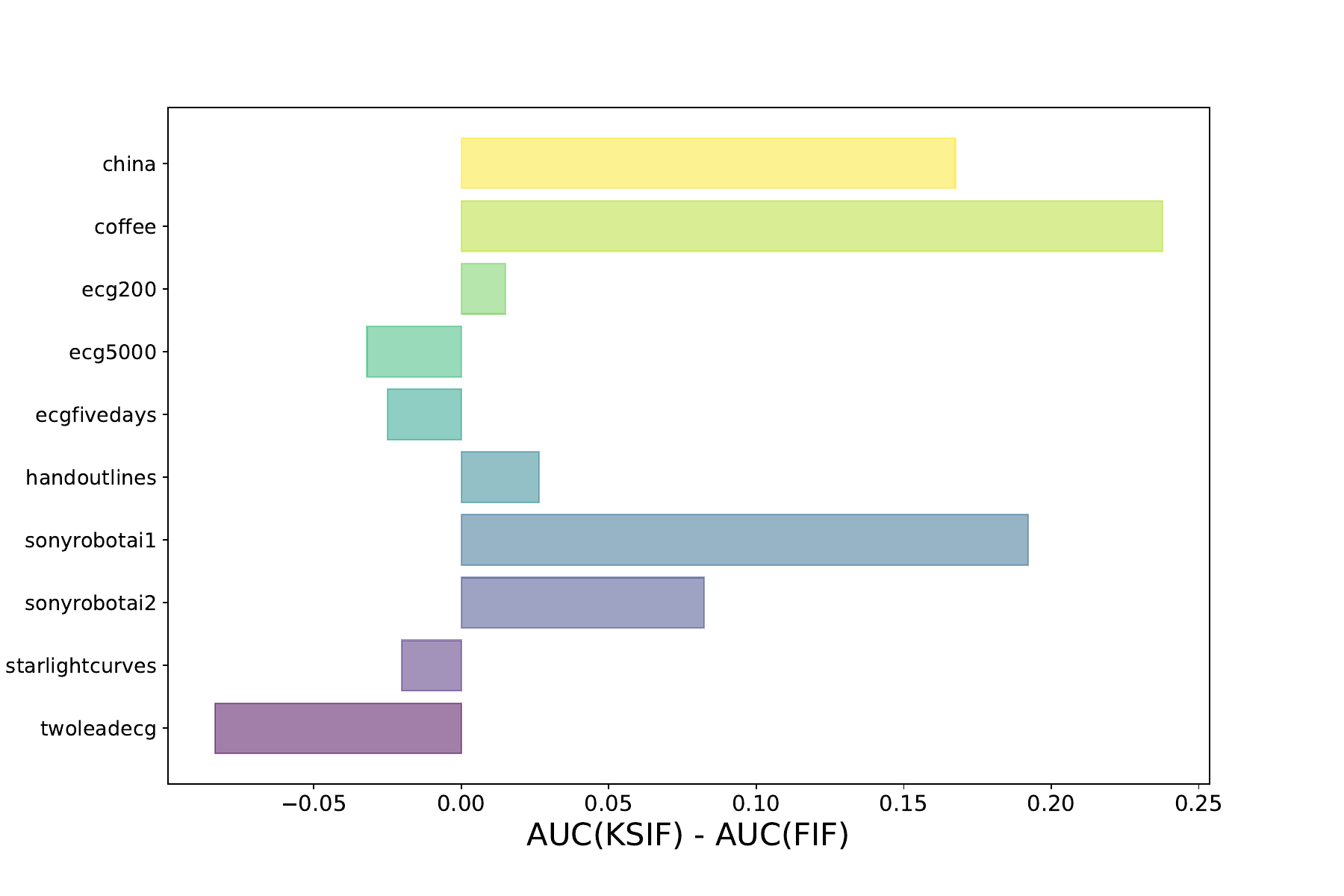} \\
         \includegraphics[trim=0cm 0 0 0,height=5cm]{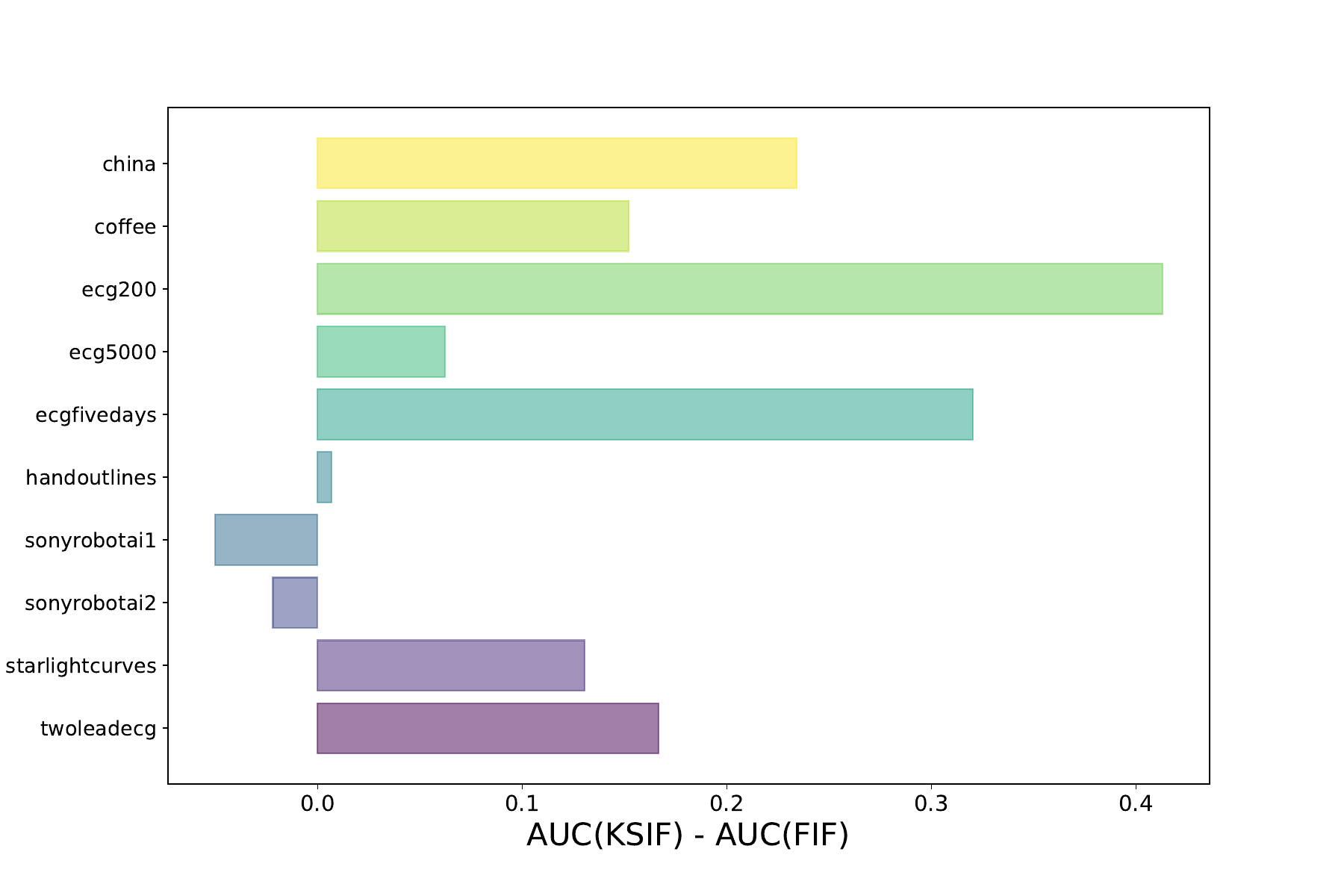}
    \end{tabular}
    \caption{Barplot of performance differences with AUC between K-SIF and FIF with a Brownian motion kernel (positive means K-SIF performs better), the inner product chose for FIF is L2 (top) and L2 of derivative (bottom).}
    \label{fig:barplot}
\end{figure}

On one hand, Figure~\ref{fig:barplot} illustrates the performance disparity between FIF and K-SIF using the Brownian dictionary. Notably, K-SIF exhibits a significant performance advantage over FIF. This observation underscores the effectiveness of the signature kernel in improving FIF's performance across most datasets, emphasizing the advantages of utilizing it over a simple inner product. On the other hand, considering the intricacy of functional data, no unique method is expected to outperform others universally. 

However, SIF demonstrates strong performance in most cases, achieving the best results for five datasets. In contrast to FIF and K-SIF, it shows robustness to the variety of datasets while not being drastically affected by the choice of the parameters involved in FIF (dictionary and inner product) and K-SIF (dictionary).

Our proposed methods, SIF and KSIF, demonstrate strong computational efficiency compared to both traditional and deep learning-based anomaly detection methods. As shown in Table~\ref{tab:computational_time}, SIF and KSIF consistently achieve lower execution times than deep learning approaches, such as AutoEncoder and AnoGan, which require significantly more computational resources. Notably, on datasets like HandOutlines and StarLightCurves, KSIF and SIF exhibit a substantial reduction in processing time compared to deep learning models, making them more suitable for real-time or resource-constrained applications. While traditional methods like IF and OCSVM offer comparable efficiency, our methods maintain a favorable balance between speed and detection performance. These results highlight the advantage of our approach in scenarios where rapid anomaly detection is critical, reinforcing its practicality for real-world deployment.

\section{Discussion \& Conclusion}\label{sec:disc-concl}

This work presents two novel anomaly detection algorithms, K-SIF and SIF, rooted in the isolation forest structure and the signature approach from rough path theory. Our contributions extend Functional Isolation Forest in two vital dimensions: incorporating non-linear properties in data for improved adaptability to challenging datasets and introducing an entirely data-driven technique free from predefined dictionaries. Such flexibility accommodates diverse data patterns and reduces the risk of overlooking certain types of anomalies. In this way, more complex real data pattern scenarios can be analysed, where non-linearity is highly present and, further, the unsupervised settings lacking labels highly affect the AD task.  We demonstrate the advantages of utilizing K-SIF over FIF through a comprehensive parameter analysis, with consistent outperformance on real-world datasets. Notably, SIF achieves state-of-the-art performance while maintaining simplicity and computational efficiency, underscoring its effectiveness in functional anomaly detection. This work offers valuable advancements in anomaly detection methodologies, providing robust solutions for complex and diverse datasets.




\bibliography{sif}
\bibliographystyle{apalike}


\clearpage
\onecolumn

\setcounter{section}{0}

\section{Additional Information About the Signature} \label{app:prop_sig}

In this section, we provide additional information regarding the signature. Firstly, the formula for the coordinate signature is provided, followed by some further information about the truncated signature and the link of the signature with moments. We then introduce references and further discussion for the computation of the signature, the linear-closed form utilised and, lastly, a review of the inner product of tensors required for the signature method is presented.

\noindent \textbf{Example of Coordinate Signature Formulas.} The definition of the signature shows that given $\bm{X}: \mathcal{F}^d \Rightarrow \mathbb{R}^d$ then these can be calculated recursively as follows
\[
S_{(i_1)}(\bm{X}) = \int_{s < u_1 < t} \mathrm{d}\bm{X}_{u_1}^{i_1} = X_{t}^{i_1} - X_s^{i_1},
\]
and 
\begin{equation*}
\begin{split}    
S_{(i_1,i_2)}(\bm{X})_{[s,t]} &= 
\int_{s < u_1 < t} S_{(i_1)}(\bm{X})_{[s,u_1]} \mathrm{d}\bm{X}_{u_1}^{i_2}\\
&= \int_{s < u_1 < u_2 < t} \mathrm{d}\bm{X}_{u_1}^{i_1}\mathrm{d}\bm{X}_{u_2}^{i_2},
\end{split}
\end{equation*}
and
\begin{equation*}
\begin{split}    
S_{(i_1,i_2,i_3)}(\bm{X})_{[s,t]} &= 
\int_{s < u_1 < t} S_{(i_1,i_2)}(\bm{X})_{[s,u_1]} \mathrm{d}\bm{X}_{u_1}^{i_3}\\
&= \int_{s < u_1 < u_2 < u_3 < t} \mathrm{d}\bm{X}_{u_1}^{i_1}\mathrm{d}\bm{X}_{u_2}^{i_2}\mathrm{d}\bm{X}_{u_3}^{i_3},
\end{split}
\end{equation*}
which gives the generic iterative $k$ iterated integral
\begin{equation*}
\begin{split}    
S_{(i_1,\ldots,i_k)}(\bm{X})_{[s,t]} &= 
\int_{s < u_1 < t} S_{(i_1,\ldots,i_{k-1})}(\bm{X})_{[s,u_1]} \mathrm{d}\bm{X}_{u_1}^{i_k}.
\end{split}
\end{equation*}

\noindent \textbf{Truncated Signature.} The number of coefficients for a multivariate function of dimension $d$ and a level of truncation $k$ is then equal to $\sum_{j=0}^{k} d^j$ if $d>1$ and equal to $k+1$ if $d=1$. It is often convenient to remove the first coefficient equal to one since it does not provide any relevant information about the function. It is worth noting that the number of coefficients increases exponentially with $k$ and polynomially with $d$. Thus, to be computed in practice, one has to choose a reasonable  depth according to the dimension of the underlying data.

\noindent \textbf{Link with Moments.} Let $\mathbf{X}$ be a stochastic process of bounded variation, authors of \cite{chevyrev2016primer} constructed a characteristic function for $X$ as $M\mapsto \mathbb{E}[M(S(\mathbf{X}))]$. They demonstrated that empirical statistical moments, such as the empirical mean using the first-order coefficients and variance using the second-order coefficient, can be explicitly recovered from signature coefficients. They also show  that if $\mathbb{E}[S(\mathbf{X})]$ is well defined, then the law of the stochastic process $\mathbf{X}$ is entirely determined by $\mathbb{E}[S(\mathbf{X})]$. Thus, the signature serves as the counterpart of an exponential of the moment-generating function in the context of vector-valued processes, and signature coefficients of order $k$ can be likened to moments.

\noindent \textbf{Computation of the Signature.} The signature is usually computed on the linear paths reconstructed from observations \citep{lyons2022signature,fermanian2021embedding} thanks to the two properties listed below.
First, the signature benefits from a closed form when computed on linear paths. Second, a concatenation formula has been provided by \cite{chen1958integration}, where the coordinate signature of two concatenate segments is given by a tensor product of the specific coordinate signatures on each segment. These two properties are described formally below.

\noindent \textbf{Linear Closed-Form.}
Let $\mathbf{X} \in \mathcal{F}^d([0, 1])$ a linear function such that $X_t =(X^1_t, \ldots , X^d_t)=(a_1+b_1 t, \ldots, a_d+b_d t)$. Then the coordinate signature on $[s, t]\subset [0,1]$ has the following closed-form:
\begin{equation}\label{eq:closed_form}
    S_{i_1,\ldots, i_k}(\mathbf{X}) = \dfrac{b_{i_1}\ldots b_{i_k} (t-s)^k}{k!}.
\end{equation}

\begin{proposition}[Chen's Identity, \citealp{chen1958integration}] Let $\mathbf{X} \in \mathcal{F}^d([s, t])$ and $\mathbf{Y}  \in \mathcal{F}^d([s, t])$ two functions with bounded variation. Then for any index $(i_1, \ldots, i_k) \subset \{1, \ldots, d \}^k$,

\begin{equation*}
    S_{(i_1, \ldots, i_k)} (\mathbf{X}  * \mathbf{Y} ) = \sum_{\ell = 0}^{k} S_{(i_1, \ldots, i_{\ell})}(\mathbf{X}) \cdot S_{(i_{\ell+1}, \ldots, i_k)}(\mathbf{Y}).
\end{equation*}

\end{proposition}

In practice, the signature is usually computed as follows. First,  a linear path is reconstructed by interpolation from the observed data. In a second step, the signature of each segment of the linear path are calculated separately thanks to the closed-form \eqref{eq:closed_form}. Then the signature of the entire path is computed recursively using the Chen's identity property.


\begin{remark}
    The signature kernel is solution of the Goursat PDE  for continuously differentiable paths \citep{salvi2021signature} and then can be approximated by a finite difference scheme. The scheme proposed by \cite{salvi2021signature} involves a computational complexity of $O(d^2 m^2 2^{2\lambda})$ where $m$ is the number of observations of the two functions and $\lambda$ is the parameter of the dyadic refinements of the grid $D \times D$. In our case, this computation occurs at each node of each tree and then the quadratic dependency in $d$ and $m$ may be computationally expensive. Therefore, we focus on computing the truncated kernel signature, that leads to a linear complexity in both the dimension $d$ and the observation number $m$.
\end{remark}

\noindent \textbf{Definition of Inner Product of Tensors for Signature Method.} Following \cite{chevyrev2022signature} and \cite{lee2023signature}, we introduce the inner product of the tensors required to compute the truncated signature and its kernel. 

Consider a vector space $V^m$ such that $\forall
m \; \exists \; \{ {\bf v}_1, \dots, {\bf v}_{d_m} \}$ basis vectors, where $d_m = | \{ 1, \dots, d \}^m  |$.  Then, any vector $S_m(\mathbf{X}) \in V^m$ given as 
\begin{equation*}
S_m(\mathbf{X}) = \left(  S_{\underbrace{(1,\ldots, 1)}_{m}} (\mathbf{X}) , \ldots,  S_{\underbrace{(d,\ldots, d)}_{m}} (\mathbf{X}) \right),
\end{equation*}
can be expressed as a weighted combination of the basis functions in $V^m$. 

Remark now that, if $V^m, V^{m'}$, are vector spaces (possibly infinite-dimensional), then there exists another vector space $V^{m} \otimes V^{m'}$ and a bilinear map $\varphi: V^{m} \times V^{m'} \rightarrow V^{m} \otimes V^{m'}$ with the universal property that any other bilinear map $V^{m} \times V^{m'} \rightarrow W$ factors uniquely through $\varphi$. This map $\otimes$ is called the tensor product, and we call the elements of $V^{m} \otimes V^{m'}$ tensors. In particular, for $m \geq 0$, we denote an element of $V^m \otimes \dots \otimes V^m = \prod_{n=1}^N (V^m)^{\otimes n}$ a tensor of degree $n$. 

If we then consider $S (\mathbf{X}) \otimes  S (\mathbf{Y})$ where $S (\mathbf{X}) \in$ $V^m$ and  $S (\mathbf{Y}) \in$ $V^{m'}$, then the so-called tensor convolution product of $S (\mathbf{X}) \otimes  S (\mathbf{Y}) \in V^m \otimes V^{m'}$, which is also a vector space with basis vectors given as
\begin{equation}\label{eq:basis_conv_space}
 \{ {\bf v}_1^m \otimes {\bf v}_1^{m'}, {\bf v}_1^m \otimes {\bf v}_2^{m'} \ldots, {\bf v}_{d_m}^m \otimes {\bf v}_{d_{m^{'}} }^{m'}   \},   
\end{equation}
$\prod_{n \geq  0} (V^m)^{\otimes n}$ represents the most general algebra containing $V^m$ \cite{reutenauer2003free}. One can also refer to $\prod_{m =  0}^{k} V^{\otimes m}$  which is again a linear
space and also forms an algebra with the tensor convolution product given in Equation \ref{eq:basis_conv_space}, restricted to the first $k$ tensors. Following this discussion, then in the notation below, $(\mathbb{R}^d)^{\otimes m}$ represents the space of $m$-tensors in $d$ dimensions. 

In practice, there is a natural inner product on $\mathbb{R}^d$ by setting
\begin{equation*}
\langle S_m^k(\mathbf{X}), S_m^k(\mathbf{Y}) \rangle_m := \sum_{(i_1, \dots, i_k) \in \{1, \ldots, d\}^k } S_{(i_1, \dots, i_k)}(\mathbf{X}) \; S_{(i_1, \dots, i_k)}(\mathbf{Y}),
\end{equation*}
where $S_m^k(\mathbf{X}), S_m^k(\mathbf{Y}) \in (\mathbb{R}^d)^{\otimes m}$ and  $m \in \{1, \dots, k \}$. Such an inner product can be extended to the subset of $\prod_{m=1}^{k} (\mathbb{R}^d)^{\otimes m}$ as
\begin{equation*}
\langle S^k (\mathbf{X}), S^k (\mathbf{Y}) \rangle = \sum_{m = 0}^{k} \langle S_m^k(\mathbf{X}), S_m^k(\mathbf{Y}) \rangle_m,
\end{equation*}
with $S^k (\mathbf{X}), S^k (\mathbf{Y}) \in \prod_{m=1}^{k} (\mathbb{R}^d)^{\otimes m}$. Note that, elements in $\prod_{m=1}^{k} (\mathbb{R}^d)^{\otimes m}$ can be tuples where each component is a tensor of degree $m$ in $d$ dimensions.

\section{K-SIF and SIF Algorithms} \label{app:algos}

This section provides the algorithms for the two proposed methods, Kernel Signature Isolation Forest and Signature Isolation Forest. The steps for each algorithm are described and presented below, introducing the input of each procedure and the steps followed to construct the nodes of the partition trees and the children subsets and datasets. Finally, the output of each method is given. The details of these procedures are provided in the main paper in Section \ref{sec:SIF}. Note that the $\omega$ parameter, corresponding to the number of split windows used for the signature, is considered as input since it must be chosen for the procedures to advance. Still, it is hidden within the presentation of the algorithms.

We further provide the following remark to explain what is the main link between the two proposed algorithms.

\begin{remark}[\textsc{Link between K-SIF and FIF}]  The first order coefficients of the signature on an interval $[s, t]\subset [0, 1]$ represent the displacement of the function:
$$S_1(\mathbf{X})=\int_{s}^{t}  \mathrm{d}X_{u_1}= X_t - X_s.$$
Therefore, additional coefficients are needed to discriminate real-world functional data. Given a finite number of observations of a function, $0\leq t_1, \ldots, t_m\leq 1$, and choosing $m$ as the number of splitting windows, the inner product between the one order signature two functions $\mathbf{X}$ and $\mathbf{Y}$ with bounded variations is defined as  $\langle S_1(\mathbf{X}), S_1(\mathbf{Y})\rangle$ is equal to the approximation of the inner product $\langle \mathbf{X}', \mathbf{Y}'\rangle_{L_2}$ used by FIF \citep{staerman2019functional}. Functional Isolation Forest with $\alpha=0$ corresponds to K-SIF with a truncated one-level. Therefore, our approach generalizes the part of FIF, considering the first moment of the underlying functions.
\end{remark}

\begin{algorithm*}[h]
\caption{\textbf{Algorithms}}
\vspace{-0.5cm} 
\footnotesize

\begin{multicols}{2}

\hrule \vspace{0.1cm}
\textbf{Kernel Signature Isolation Tree} \vspace{0.1cm}
\hrule

\vspace{0.1cm}

\begin{algorithmic} 
\footnotesize

\INPUT   A subsample $\{ {\bf x}_{i_1}, \dots, {\bf x}_{i_m} \}$, a dictionary $\mathcal{D}$, a probability measure $\boldsymbol{\nu}$,  the truncated level of the signature $k$,  the number of splitting windows $\omega$

\vspace{0.4cm}

\STATE \textbf{(a)} The root node indexed by $(0,0)$ is associated with the whole input space
$\mathcal{C}_{0,0} = \mathcal{F}^d$. 

\STATE \textbf{(b)} If the node $(p,q)$ is terminal, stop the construction, otherwise go to  \textbf{(c)}.

\vspace{0.3cm}

\STATE \textbf{(c)} A non-terminal node $(p,q)$ is split as follows:

\begin{enumerate}
    \item Choose a split variable ${\bf d}$ according to the probability distribution $\boldsymbol{\nu}$ on $\mathcal{D}$.
    
    \item Choose randomly and uniformly a split value $\gamma$ in the interval
    
    {\footnotesize
    \begin{equation*}
     \hspace{-0.9cm}\left[ \min_{ {\bf x} \in \mathcal{X}_{p,q} } \langle S^k ({\bf x}), S^k ({\bf d}) \rangle  , \max_{ {\bf x} \in \mathcal{X}_{p,q} } \langle S^k ({\bf x}), S^k ( {\bf d} ) \rangle   \right],   
    \end{equation*}}
    where $S^k ({\bf d})$ is the signature of ${\bf d}$.
    
    \item Form the children subsets
    \begin{equation*}
        \begin{split}
 \mathcal{C}_{p+1,2q} &= \mathcal{C}_{p,q} \cap C_{\text{K-SIF}}^L \\
\mathcal{C}_{p+1,2q+1} &= \mathcal{C}_{p,q} \cap C_{\text{K-SIF}}^R,
        \end{split}
    \end{equation*}
    as well as the children training datasets
    \begin{equation*}
    \begin{split}
            \mathcal{X}_{p+1,2q} &= \mathcal{X}_{p,q} \cap \; \mathcal{C}_{p+1,2q} \\
            \mathcal{X}_{p+1,2q+1} &= \mathcal{X}_{p,q} \cap \; \mathcal{C}_{p+1,2q+1}.
                \end{split}
     \end{equation*}
\end{enumerate}

\vspace{0.3cm}

\STATE \textbf{(d)} Apply the building procedure starting from \textbf{(b)} to nodes $(p + 1, 2q)$ and $(p + 1, 2q + 1)$

\vspace{0.3cm}

\OUTPUT  $(\mathcal{C}_{0,0}, \mathcal{C}_{1,1}, \dots,  )$

\end{algorithmic}

    \columnbreak 
 
\hrule \vspace{0.1cm}
\textbf{Signature Isolation Tree} \vspace{0.1cm}
\hrule

\vspace{0.1cm}

\begin{algorithmic}
\footnotesize

\INPUT  A subsample $\{ {\bf x}_1, \dots, {\bf x}_{m} \}$, $k$ depth of the signature, the truncated level of the signature $k$,  the number of splitting windows $\omega$

\vspace{0.7cm}

\STATE \textbf{(a)}  The root node indexed by $(0,0)$ is associated with the whole input space
$\mathcal{C}_{0,0} = \mathcal{F}^d$. 

\STATE \textbf{(b)} If the node $(p,q)$ is terminal, stop the construction, otherwise go to  \textbf{(c)}.

\vspace{0.3cm}

\STATE  \textbf{(c)} A non-terminal node $(p,q)$ is split as follows:

\begin{enumerate}
    \item Choose randomly  and uniformly a coordinate $(i_1,\ldots, i_\ell)$ in the set 
$$\{ (i_1, \ldots
, i_\ell) \in \llbracket  1, d \rrbracket^\ell; \quad 1\leq \ell \leq k \}.$$

    \item Choose randomly and uniformly a split value $\gamma$ in the interval
    {\footnotesize
    \begin{equation*}
     \hspace{-0.6cm} \left[ \min_{ {\bf x} \in \mathcal{X}_{p,q} } S_{(i_1,\ldots, i_\ell)}(\mathbf{x})  , \; \; \max_{ {\bf x} \in \mathcal{X}_{p,q} } S_{(i_1,\ldots, i_\ell)}(\mathbf{x})   \right].
    \end{equation*}    }
     \item Form the children subsets
    \begin{equation*}
        \begin{split}
 \mathcal{C}_{p+1,2q} &= \mathcal{C}_{p,q} \cap C_{\text{SIF}}^L \\
\mathcal{C}_{p+1,2q+1} &= \mathcal{C}_{p,q} \cap C_{\text{SIF}}^R,
        \end{split}
    \end{equation*}
    as well as the children training datasets
    \begin{equation*}
    \begin{split}
            \mathcal{X}_{p+1,2q} &= \mathcal{X}_{p,q} \cap \; \mathcal{C}_{p+1,2q} \\
            \mathcal{X}_{p+1,2q+1} &= \mathcal{X}_{p,q} \cap \; \mathcal{C}_{p+1,2q+1}.
                \end{split}
     \end{equation*}
\end{enumerate}

\vspace{0.3cm}

\STATE \textbf{(d)} Apply the building procedure starting from \textbf{(b)} to nodes $(p + 1, 2q)$ and $(p + 1, 2q + 1)$

\vspace{0.3cm}

\OUTPUT  $(\mathcal{C}_{0,0}, \mathcal{C}_{1,1}, \dots,  )$

\end{algorithmic}
\end{multicols}
\end{algorithm*}

\section{Additional Numerical Experiments} \label{app:add_experiments}

In this section, we present additional numerical experiments in support of the proposed algorithms and arguments developed in the main body of the paper. First, We describe the signature depth's role in the algorithms and explain how this parameter affects them. We provide boxplots for two sets of generated data and argue the importance of the depth parameter in this context. Afterwards, we provide additional experiments on the robustness to noise advantage of  (K)-SIF over FIF, related to Section 4 of the main body of the paper. The third paragraph refers to the generated data for the 'swapping events' experiment in Section 4.2 of the main body of the paper is shown. We provide a Figure for visualization and a better understanding. We further remark on how we constructed the data. The fourth subsection then demonstrates the computational time of the proposed algorithms with a direct comparison to FIF. Then, an additional experiment presenting further evidence for the discrimination power with respect to the AD task of (K)-SIF over FIF is presented. Finally, the last subsection shows a Table which describes information about the size of datasets related to the benchmark in Section 4.3.

\subsection{The Role of the Signature Depth} \label{subsec:depth}

All moments of stochastic processes with bounded variations can be characterized by the signature \citep{chevyrev2016primer}. Following this rationale, the truncated signature encapsulates more information with higher truncation levels. However, increasing the truncated depth comes at the cost of longer computational times, given that the number of signature coefficients scales with $O(d^k)$, which can be restrictive in high-dimensional settings. Therefore, the choice of the truncation level becomes a trade-off between computational efficiency and the information retained by the signature.

In this experiment, we investigate the impact of this parameter on K-SIF with two different classes of stochastic processes. The three-dimensional Brownian motion (with $\mu = 0$ and $\sigma = 0.1$), characterized by its two first moments, and the one-dimensional Merton-jump diffusion process, a heavy-tail process widely used to model the stock market. In such a way, we compare the former class of stochastic models to the latter, which, instead, cannot be characterized by the first two moments and observe performances of (K)-SIF in this regard.

We computed K-SIF with three dictionaries with truncation levels varying in $\{2, 3, 4\}$ for both simulated datasets. We set the number of split windows to 10, according to the previous section, and the number of trees to $1000$. After that, we computed the Kendall correlation of the rank returned by these models for the three pairwise settings: level 2 vs level 3, level 2 vs level 4, and level 3 vs level 4. 

\begin{figure}[!ht]
    \centering
    \begin{tabular}{cc}
       \includegraphics[trim=0cm 0 0 0,height=4.5cm]{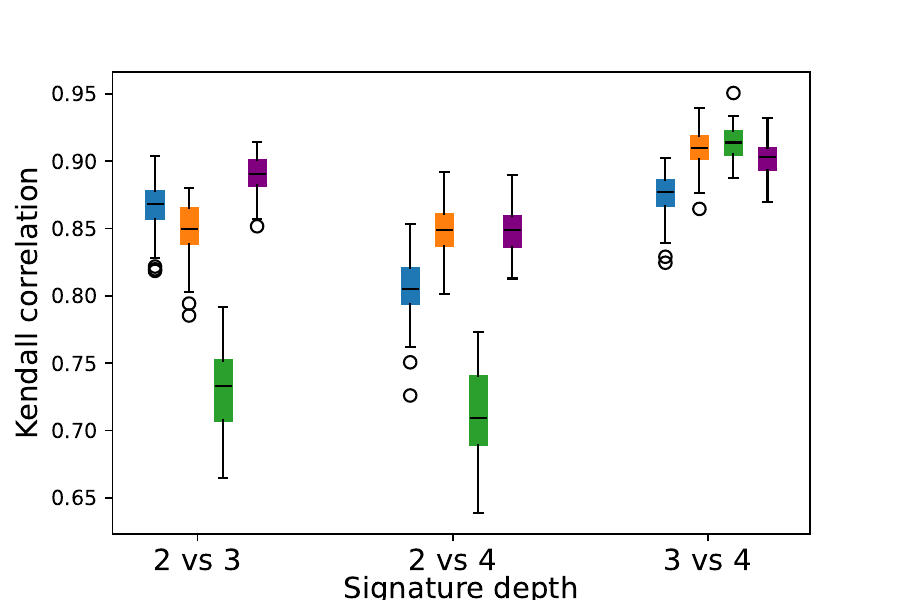}    &  \includegraphics[trim=0cm 0 0 0,height=4.5cm]{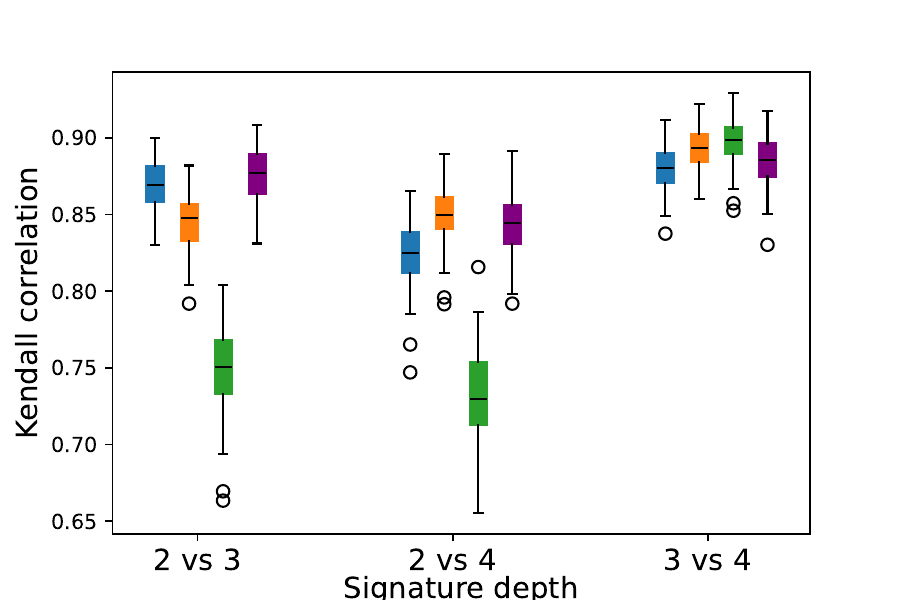} 
    \end{tabular}

    \caption{\textbf{Brownian Motion Process Results.} Kendall tau correlation between the score returned by SIF (purple) and K-SIF with different depth values,  $\omega=3$ (left) and $\omega=5$ (right) , for the three dictionaries: ‘Brownian' (blue), ‘Cosine' (orange) and ‘Gaussian wavelets' (green) on three dimensional Brownian paths.}
    \label{fig:depth}
\end{figure}

\begin{figure}[!ht]
    \centering
\begin{tabular}{cc}
    \includegraphics[trim=0cm 0 0 0,height=4.5cm]{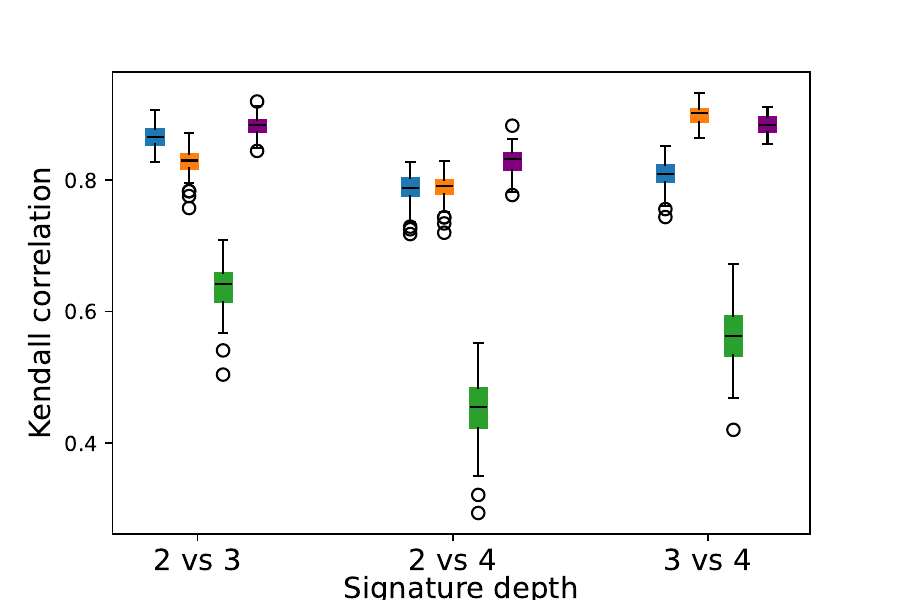} & 
    \includegraphics[trim=0cm 0 0 0,height=4.5cm]{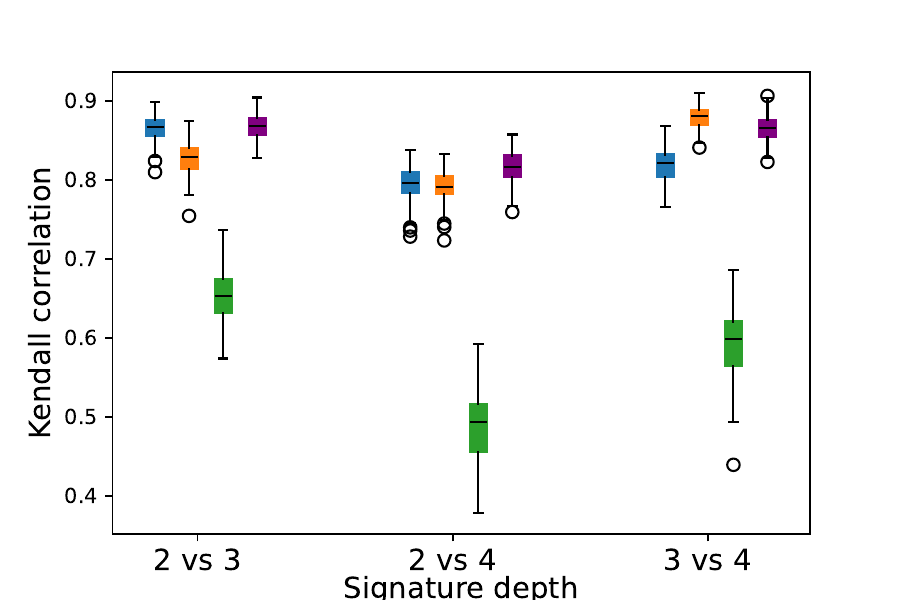}
\end{tabular}
    \caption{\textbf{Merton-Jump Diffusion Process Results.} Kendall tau correlation between the score returned by SIF (purple) and K-SIF with different depth values for the three dictionaries: ‘Brownian' (blue), ‘Cosine' (orange) and ‘Gaussian wavelets' (green) with $\omega=3$ (left) and $\omega=5$ (right) on Merton-jump diffusion processes.}
    \label{fig:depth_merton}
\end{figure}

We repeated this experiment 100 times and report the correlation boxplots in Figure~\ref{fig:depth} for the Brownian motion and in Figure~\ref{fig:depth_merton} for the Merton-jump diffusion process. Note that the left and right plots refer to the different split window parameters selected for K-SIF, corresponding to $\omega = 3$ for the left panels, while, for the right ones, we chose $\omega = 5$. These boxplots show the Kendall tau correlation between the score returned by one of the algorithms used with one specific depth and the same algorithm with a different depth. K-SIF results with the three dictionaries are represented in blue, orange, and green for the Brownian, Cosine and green Gaussian wavelets, respectively. SIF boxplots are instead in purple. The y-axis refers to the Kendall correlation values and the x-axis to the settings of the depth values with respect to which the correlation has been.

A high correlation indicates an equivalent rank returned by the algorithm with different depth parameters. Therefore, if the correlation is high, this suggests that this parameter does not affect the results of the considered algorithm, and a lower depth should be selected for better computation efficiency. High correlations are shown for both SIF (purple boxplots) and K-SIF for the two dictionaries, i.e. Brownian and Cosine (blue and orange boxplots). Therefore, choosing the minimum truncation level is recommended to improve computational efficiency. For the same algorithms, slightly lower correlations are identified in the case of the Merton processes, yet still around 0.8 levels, hence supporting an equivalent claim. In the case of K-SIF with the Gaussian dictionary (green boxplots), a much higher variation is obtained regarding correlation results across the three tested scenarios. Furthermore, in the case of the Merton-jump diffusion processes, the results show a lower correlation, consistent with the other results. Therefore, in the case of K-SIF with such a dictionary, the depth should be carefully chosen since different parameters might lead to better detection of the moments of the underlying process.

\subsection{Robustness to Noise }
 We now explore the sensitivity of our class of algorithms to noisy data. We also provide a comparison with Functional Isolation Forest. To that end, we simulate a dataset of 500 ‘relatively smooth' standard Brownian motion paths with $\mu = 0$ and $\sigma = 0.05$. We also simulate 50 standard Brownian motion paths with $\mu = 0$ and $\sigma \in [0.05, 0.2]$. Depending on the level of noise present in the last 50 curves (i.e., standard deviation here), these paths may represent normal, noisy and abnormal data. For example, $0.05$ (corresponding to $0$ noise level) is normal data, while $0.2$ (corresponding to $0.15$ noise level) is abnormal.

\begin{figure}[!ht]
    \centering
    \includegraphics[trim=0cm 0 0 0,height=4cm]{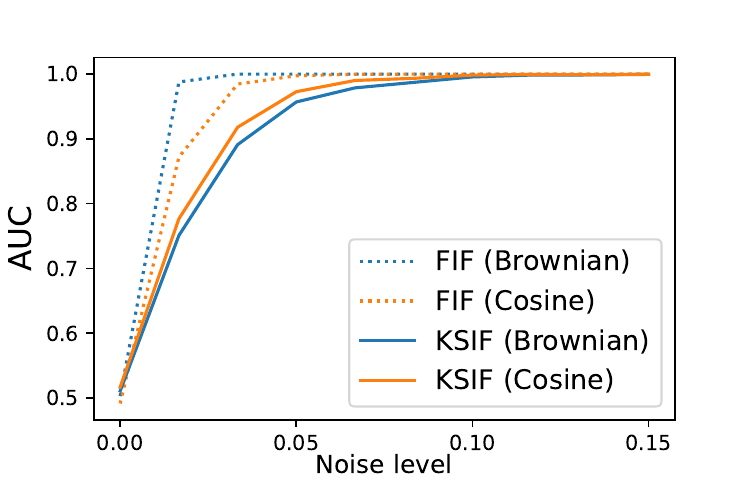}
    \caption{Robustness of K-SIF and FIF to noise level.}
    \label{fig:rob_noise}
\end{figure}

We perform K-SIF and FIF with both ‘Brownian' and ‘Cosine' dictionaries and report the AUC under the ROC curves by considering the 50 curves as the second class in Figure~\ref{fig:rob_noise}. 

An AUC of 1 means that the 50 curves of the second class are identified as anomalies, while an AUC of 0.5 indicates that the algorithm cannot distinguish them from normal data. According to the findings provided in Section~\ref{sub:parameters}, the parameters for K-SIF have been chosen such that the number of split window sizes is equal to 10 while the depth of the signature $k=2$. We can observe that FIF, regardless of the dictionary used, is highly sensitive to noisy data and considers it abnormal, even with a minimal noise level. In contrast, K-SIF is way more robust to such data and requires higher noise levels to start viewing this data as anomalies.

Further, we provide additional experiments on the robustness to noise advantage of (K)-SIF over FIF. The configuration for data simulation goes as follows. We define a synthetic dataset of $100$ smooth functions given by
$$\mathbf{x}(t)=30 t^q(1-t)^q, $$
with $t\in [0, 1]$ and $q$ equispaced in $[1, 1.4]$. Then, we simulate the occurrences of events by adding Gaussian noise on different portions of the functions. We randomly select 10\% of them and define abnormal new curves by adding Gaussian values on a sub-interval, i.e., $$\mathbf{x}(t)=30 t^q(1-t)^q + \varepsilon (t) \mathbb{I}(t\in [0.2, 0.5]),$$
where  $\varepsilon(t) \sim \mathcal{N}(0, 0.5)$. We select randomly 10\% again and create slightly noisy curves by adding small noise on another sub-interval compared to the first one, i.e., $$\mathbf{x}(t)=30 t^q(1-t)^q + \varepsilon (t) \mathbb{I}(t\in [0.7, 0.9]),$$
where  $\varepsilon (t) \sim \mathcal{N}(0, 0.1)$. 

\begin{figure}[!ht]
    \centering
    \begin{tabular}{cc}
     &  K-SIF \\ 
         \includegraphics[trim=0cm 0 0 0,height=3.5cm]{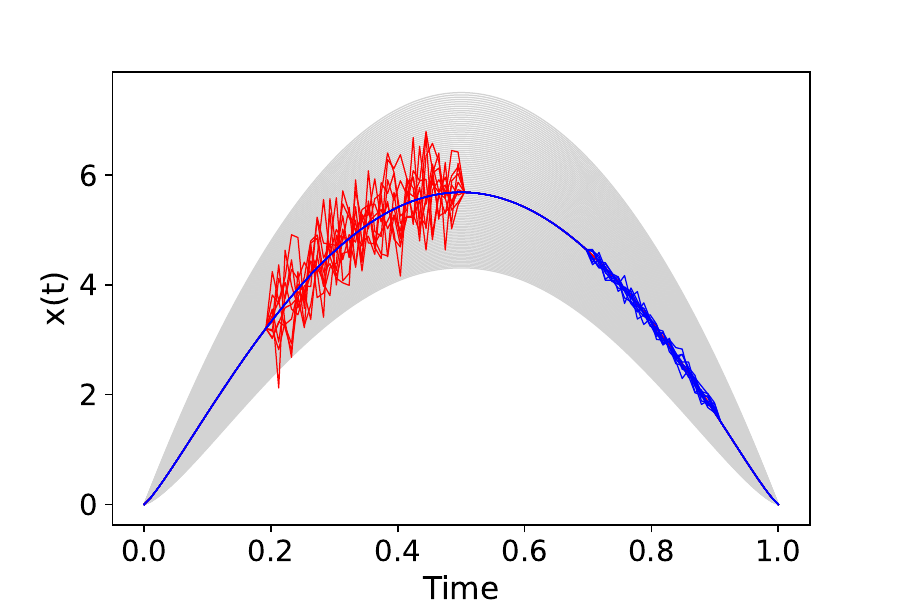} & \includegraphics[trim=0cm 0 0 0,height=3.5cm]{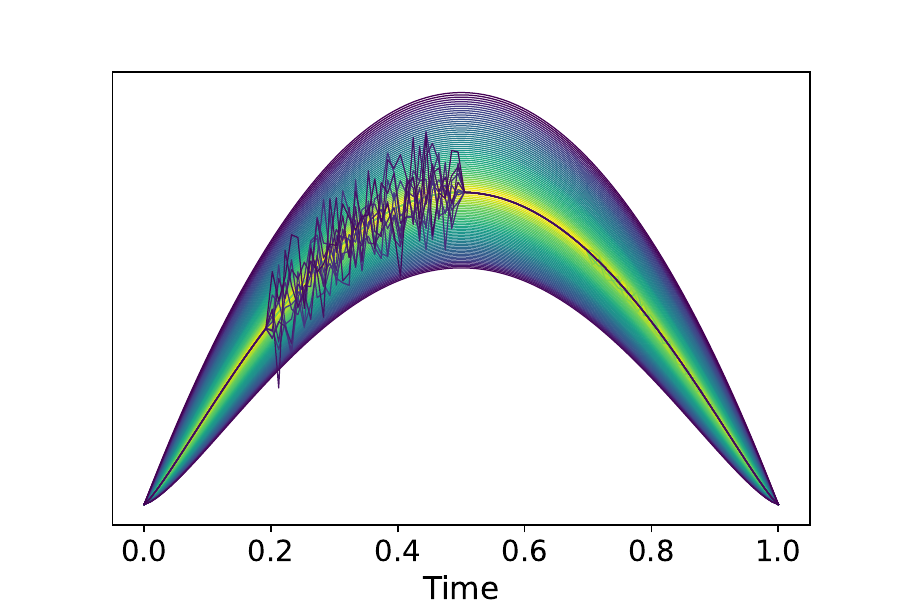} \vspace{0.5cm} \\ 
          FIF$_{1}$ &  FIF$_{0}$ \\
         \includegraphics[trim=0cm 0 0 0,height=3.5cm]{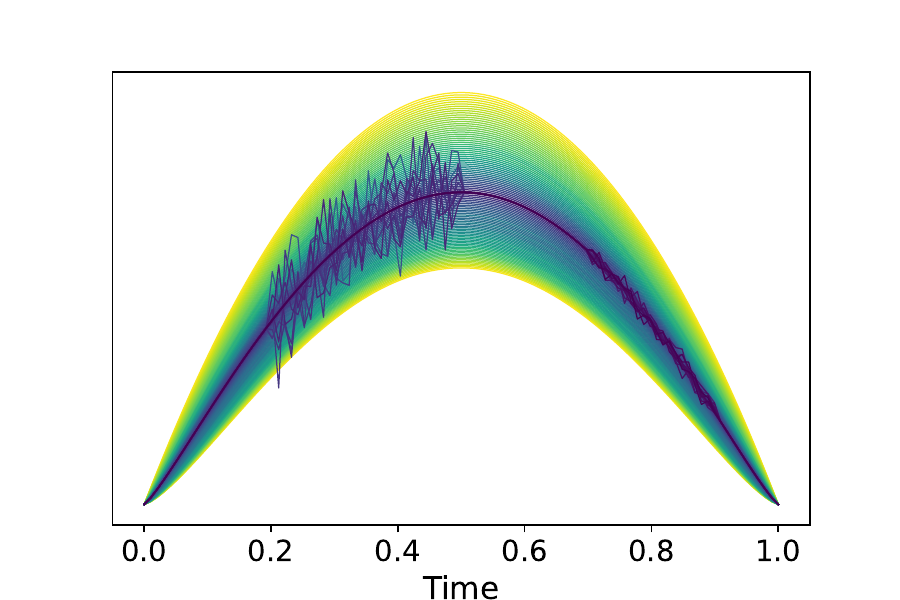}  & \includegraphics[trim=0cm 0 0 0,height=3.5cm]{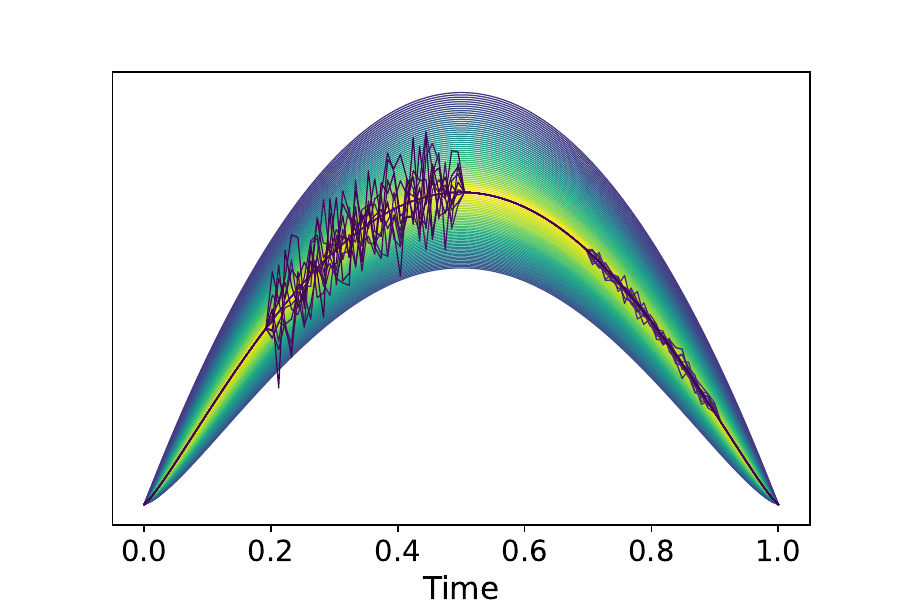} 
    \end{tabular}
    \caption{\textbf{Robustness to Noise}. The first panel presents the raw data, where there are 120 curves, of which, in red we have the 10 curves for abnormal, or noisy data, in blue the 10 curves of slightly noisy but normal data and in gray the remaining curves. The configuration for the data simulation is provided at the beginning of this section. The second, third and fourth panels show the anomaly scores assigned to the curves based on the algorithm of interest. The second panel refer to K-SIF, run with a Brownian dictionary, $k=2$ and $\omega=10$. The third and fourth panels refer to FIF run with a Brownian dictionary  with $\alpha=1$ (third) and $\alpha=0$ (fourth), respectively. The anomaly score color increases from yellow to dark blue in the second and fourth plots, i.e. a dark curve is abnormal and yellow is normal. In the third plot,  for vizualisation purposes, it decreases, i.e. a dark curve is normal and yellow is abnormal.}
    \label{fig:rob_illu}
\end{figure}

Figure~\ref{fig:rob_illu} provides a summary visualization of the generated dataset in the first panel. The 10 anomalous curves are plotted in red, while the 10 considered slightly noisy normal data are plotted in blue. The rest of the curves, considered normal data, is provided in gray. The idea is to understand how the dictionary choice influences K-SIF and FIF in detection of slightly  noisy normal data versus abnormal noise. Results for K-SIF and FIF are provided in the second, third and fourth panels of Figure~\ref{fig:rob_illu}, respectively.   

We compute K-SIF with a Brownian dictionary, $k=2$ and $\omega=10$ and FIF for $\alpha=0$ and $\alpha=1$ also with a Brownian dictionary. The colors of the panels represent the anomaly score assigned to each curve for that specific algorithm. In the second (K-SIF) and last (FIF with $\alpha = 0$) panels, the anomaly score increases from yellow to dark blue, i.e. a dark curve is abnormal and yellow is normal, while, in the third plot (FIF  with $\alpha = 1$) it is the opposite, i.e. a dark curve is normal and yellow is abnormal.

It is possible to observe how K-SIF successfully can identify noisy and abnormal data as such. Indeed, while the abnormal data are colored in dark blue, the noisy ones display a yellow color score. Instead, in FIF with $\alpha=1$ (third panel) both the abnormal and the slightly noisy curves are identified as normal data (given the reversed scale and having dark blue colors). When it comes to FIF with $\alpha=0$ (last and fourth panel), both abnormal and noisy data are scored as abnormal curves. Hence, FIF with both settings of the $\alpha$ parameter, cannot provide a different score to noise and slightly noisy data. K-SIF, instead, successfully perform such a task.

\subsection{Swapping Events Dataset}

\begin{figure}[!ht]
    \centering
         \includegraphics[trim=0cm 0 0 0,height=4.5cm]{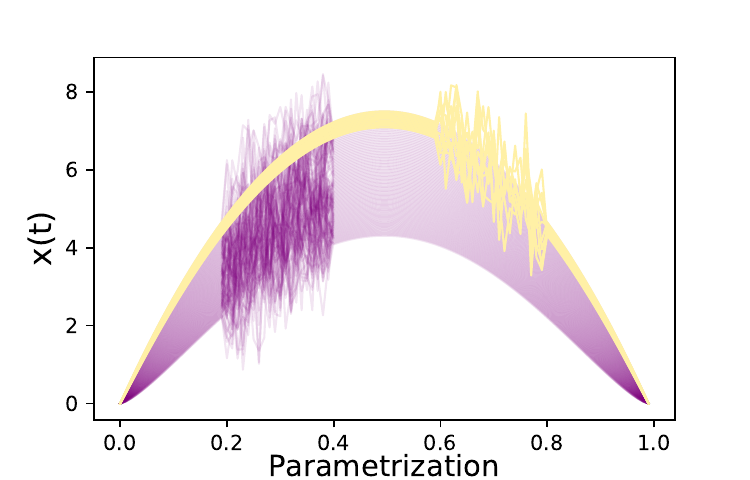}
    \caption{ \textbf{Swapping Events}. Dataset used in the experiment of Section 4.2. Purple curves represent normal data while yellow curves represent abnormal data. Configuration of the simulation are provided at the beginning of the section.}
    \label{fig:swap:data}
\end{figure}

This part provides a visualization of the dataset used in the ‘swapping events' experiment in section~\ref{subsec:advantages} of the core paper. Figure~\ref{fig:swap:data} shows the simulated data. Remark that we define a synthetic dataset of $100$ smooth functions given by
$$\mathbf{x}(t)=30 t^q(1-t)^q, $$
with $t\in [0, 1]$ and $q$ equispaced in $[1, 1.4]$. Then, we simulate the occurrences of events by adding Gaussian noise on different portions of the functions. We randomly select 90\% of them and add Gaussian values on a sub-interval, i.e., $$\mathbf{x}(t)=30 t^q(1-t)^q + \varepsilon (t) \mathbb{I}(t\in [0.2, 0.4]),$$
where  $\varepsilon(t) \sim \mathcal{N}(0, 0.8)$. We consider the 10\% remaining as abnormal by adding the same ‘events' on another sub-interval compared to the first one, i.e., $$\mathbf{x}(t)=30 t^q(1-t)^q + \varepsilon (t) \mathbb{I}(t\in [0.6, 0.8]),$$
where  $\varepsilon (t) \sim \mathcal{N}(0, 0.8)$. We then have constructed two identical events occurring at different parts of the functions, leading to \textit{isolating} anomalies. 

\subsection{Computational Time of K-SIF, SIF and FIF}

\begin{figure}[!ht]
    \centering
    \begin{tabular}{ccc}
         \includegraphics[trim=1cm 0 0 0,height=3.3cm]{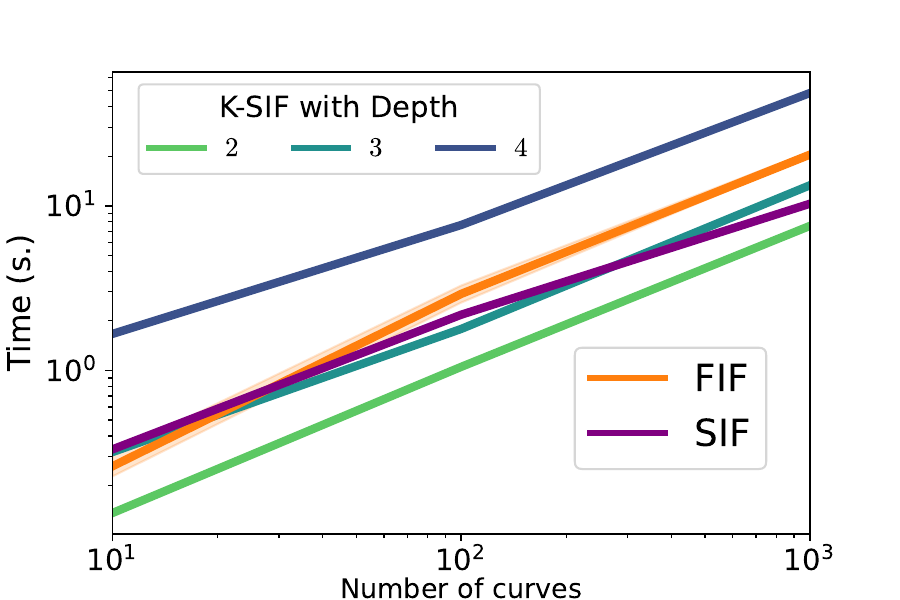} & \includegraphics[trim=2cm 0 0 0,height=3.3cm]{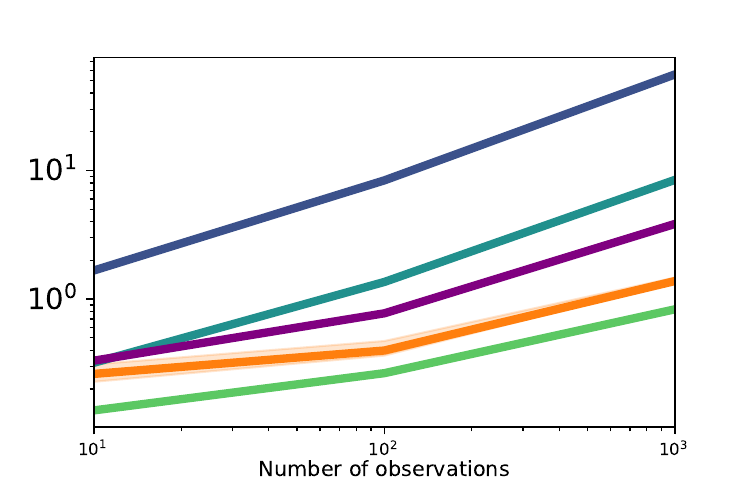}  & \includegraphics[trim=2cm 0 0 0,height=3.3cm]{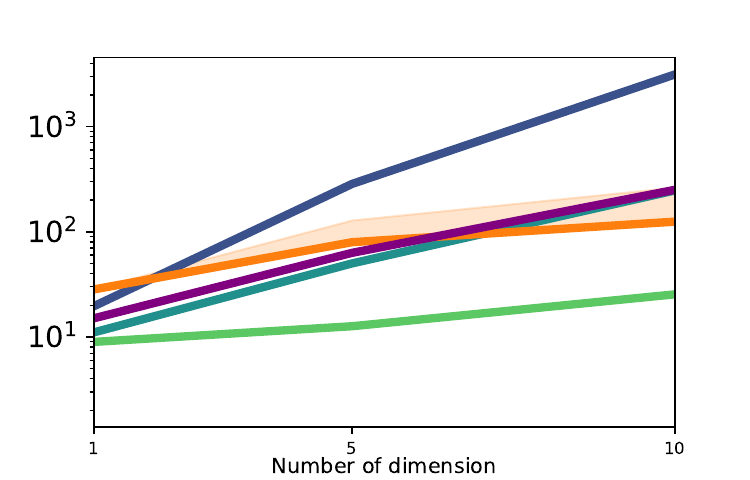} 
    \end{tabular}
    \caption{Computational time for K-SIF and FIF with respect to the number of curves (left), the number of discretization points (middle) and the number of dimension (right).}
    \label{fig:comp_time}
\end{figure}

This section investigates the computational time of K-SIF, SIF and FIF. In this experiment we simulate multidimensional Brownian motions with the dimension varying in $\{1, 5, 10 \}$, the number of curves in $\{10, 100, 1000\}$ and the number of  observation of each stochastic path in $\{10, 100, 1000\}$. The computation configuration of the three algorithms is given as follows. We compute (1) K-SIF with a Brownian motion as kernel, $\omega=10$ and a truncation level $k$ varying in $\{2, 3, 4 \}$, (2) FIF (with $\alpha=1$) and (3) SIF (with $\omega=10$, $k=3$). All methods are computed with the same tree parameters being the number of trees $N=100$, the subsample size $m = \text{min}(256,n)$ and the height limit set to $\lceil \log_2 (m) \rceil$. The experiment is repeated 10 times, and the average computation time is reported in
 Figure~\ref{fig:comp_time} w.r.t. the three parameters, $n,m,d$ of the data. We can see that K-SIF, SIF and FIF evolve comparably and that the scaling factors are equivalent. 

\subsection{K-SIF and SIF: a Better Discrimination of Anomalies compared to FIF}

In this part, we construct an additional toy experiment to show the discrimination power of (K-)SIF over FIF. We simulate 100 planar Brownian motion paths with 90\% of normal data with drift $\mu=[0, 0]$ and standard deviation $\sigma=[0.1, 0.1]$, and 10\% of abnormal data with drift $\mu=[0, 0]$ and standard deviation $\sigma=[0.4, 0.4]$.

\begin{figure}
    \centering
    \includegraphics[trim=0cm 0 0 0,height=5cm]{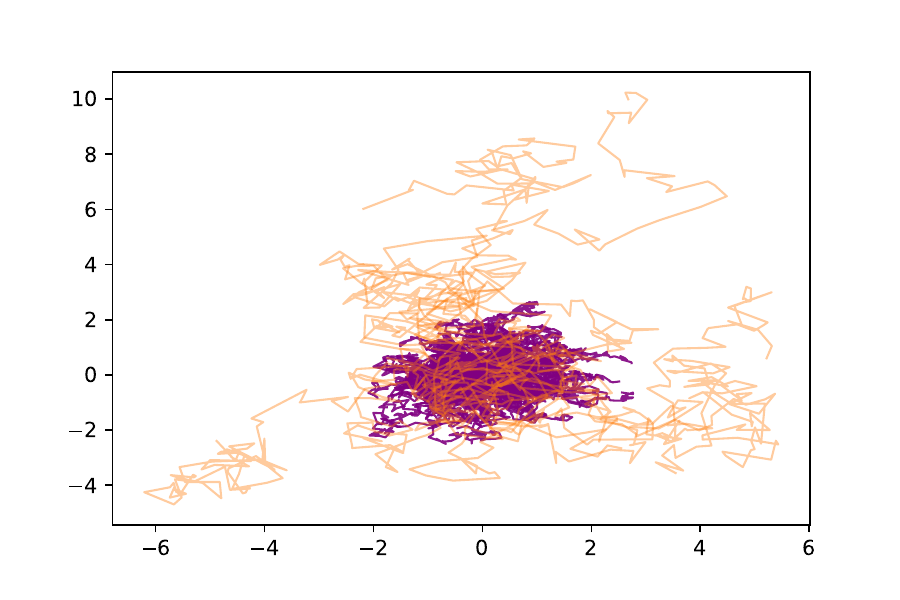}
    \caption{Dataset used for the experiment. Purple paths are normal data while orange paths are the abnormal ones. }
    \label{fig:anom_appendix}
\end{figure}

\begin{figure}[!ht]
    \centering
    \begin{tabular}{ccc}
    \hspace{-0.5cm} FIF &  \hspace{-0.5cm} K-SIF & \hspace{-0.5cm}SIF \\
         \includegraphics[trim=1cm 0 0 0,height=3.3cm]{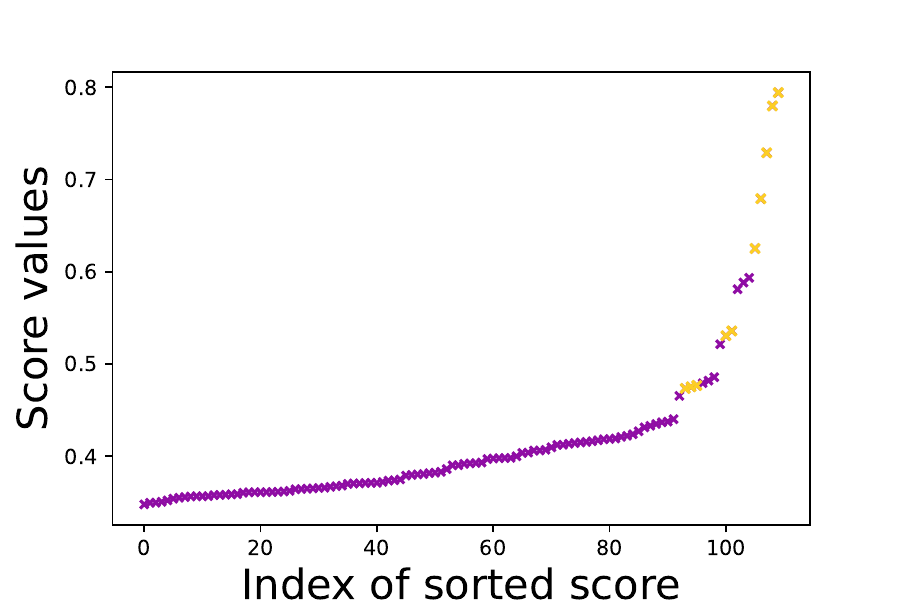} & \includegraphics[trim=2cm 0 0 0,height=3.3cm]{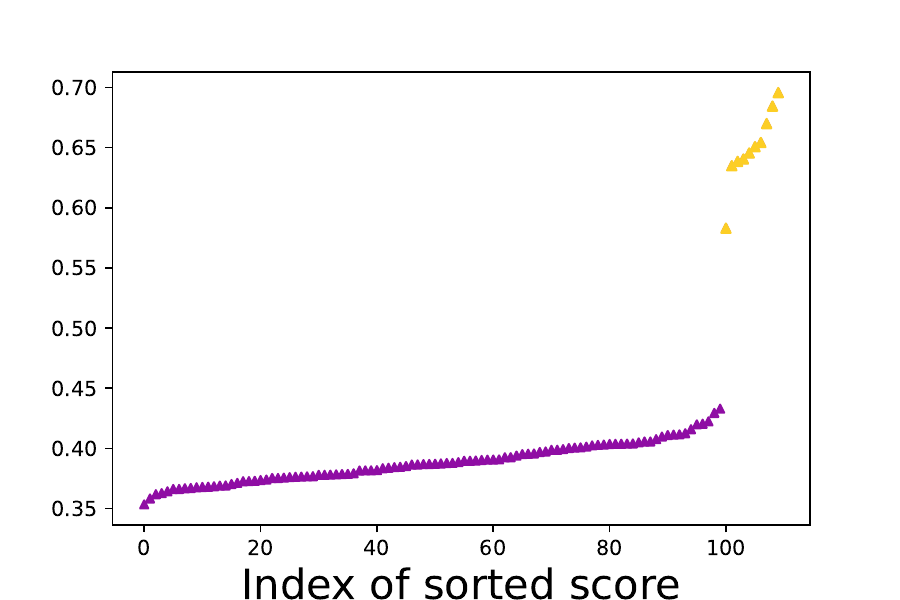}  & \includegraphics[trim=2cm 0 0 0,height=3.3cm]{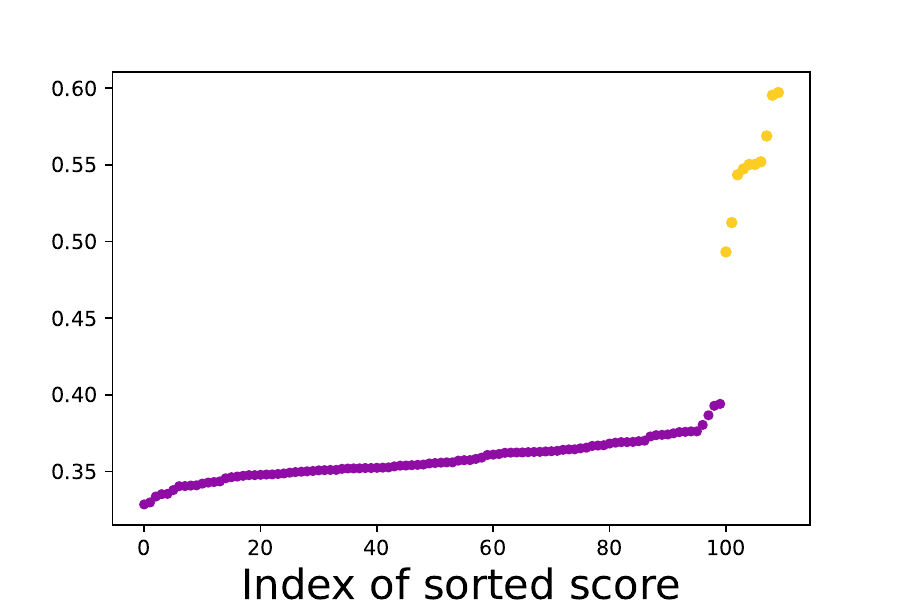} 
    \end{tabular}
    \caption{Scores returned by FIF (left), K-SIF (middle) and SIF (right) on planar Brownian motion with abnormal data (orange).}
    \label{fig:anom}
\end{figure}

Figure~\ref{fig:anom_appendix} presents one simulation of this dataset. Note that, the purple paths represent normal data, while, in orange, the abnormal are instead represented.  On this dataset, we compute FIF (with $\alpha=1$ and Brownian dictionary), K-SIF (with $k=2$, $\omega=10$ and Brownian dictionary) and SIF (with $k=2$ and $\omega=10$). To display the scores returned by the algorithsm, we provide Figure~\ref{fig:anom}. Note that, the plots shows the scores for these 100 paths, after having sorted them. Hence, the x-axis provides the index of the ordered scores, while, the y-axis represents the score values. As for the simulation, we plot in purple the scores of the normal data and in orange the scores of the abnormal data. The three panels refer to FIF, K-SIF and SIF, respectively.

It is possible to observe that the scores of K-SIF and SIF well separate the abnormal and the normal data, with a jump in the scores which is quite pronounced, i.e. the scores of the normal data are relatively distant from the scores of the abnormal data. If one focuses on FIF instead, then the discrimination of such anomalies appears to be more challenging; the first panel shows, in fact, a continuous in terms of the score returned by the AD algorithm, which does not separate normal and abnormal data.

In summary, the proposed algorithms leveraging the signature kernel (K-SIF) and the signature coordinate (SIF) exhibit more reliable results in this experimental setting, suggesting their efficacy in discerning anomalies within the simulated dataset. Detecting the order in which events happen is a much more informative feature than incorporating a functional aspect in the anomaly detection algorithm. This aspect must be further investigated and explored, particularly in the application areas where sequential data, such as time series, are taken into account.

\subsection{Anomaly Detection Benchmark Data}

This subsection presents Table~\ref{tab:datsets}, where further information about the size of datasets related to the benchmark in Section 4.3 is provided. We consider ten datasets. The columns offer $p$, corresponding to the number of discretization points, the $n_a/n$ ratios, where $n_a$ correspond to the number of anomalies and $n$ the number of samples and, lastly, the labels denoting normal and abnormal data. In addition, the FPR at 95\% FPR is given in Table~\ref{tab:aupr} and the AUPR  is given in Table~\ref{tab:fpr95tpr}. Tables \ref{tab:deep:auc}, \ref{tab:deep:aupr} and \ref{tab:deep:fpr} depict additional deep learning based methods on the same benchmark.

{\renewcommand{\arraystretch}{2} 
{\setlength{\tabcolsep}{0.2cm}
\begin{table*}[!htt]
\begin{center}
{\scriptsize
\begin{tabular}{|c||c|c|c|c|c|c|}
\hline
 & p & training/testing : $n_a$/ $n$ &normal lab & anomaly lab \\
\hline
\hline
Chinatown & 24& 4 / 14 (29\%)  &   2& 1\\
\hline
Coffee & 286& 5 / 19 (26\%) &  1 & 0\\
\hline
ECGFiveDays & 136& 2 / 16 (12\%)&  1&  2  \\
\hline
ECG200 & 96&31 / 100 (31\%) &   1&  -1 \\
\hline
Handoutlines &2709 & 362 / 1000 (36 \%) &   1& 0\\
\hline
SonyRobotAI1& 70 & 6 / 20 (30 \%) &  2 & 1\\
\hline
SonyRobotAI2 & 65 & 4 / 20 (20 \%) &  2 &  1\\
\hline
StarLightCurves & 1024 & 100 / 673 (15 \%) &  3 &  1 and 2\\
\hline
TwoLeadECG &82& 2 / 14 (14 \%) &  1&  2\\
\hline
ECG5000 & 140 & 31 / 323 (10 \%) &  1 &  3,4 and 5  \\
\hline
\end{tabular}
}
\end{center}
\caption{Datasets considered in performance comparison: $n$ is the number of instances, $n_a$ is the number of anomalies. $p$ is the number of discretization points.}
\label{tab:datsets}
\end{table*}}}

{\renewcommand{\arraystretch}{1.5} 
{\setlength{\tabcolsep}{0.18cm}
\begin{table}[!htt]
\centering
{\scriptsize
\begin{tabular}{|l|c|c|c|c|c|c|c|c|c|c|c|}
\hline
& SIF & KSIF$_{GW}$ & FIF$_{GW}$ & KSIF$_{C}$ & FIF$_{C}$ & KSIF$_{B}$ & FIF$_{B}$ & IF & OCSVM & fHD & fSDO  \\
\hline
Chinatown & \textbf{0} & 0.21 & 0.93 & 0.07 & 0.86 & \textbf{0} & 0.93 & 0.93 & 1. & 1. & \textbf{0}  \\
\hline
Coffee & 1. & \textbf{0.21} & 0.95 & 0.95 & 0.37 & 1. & 1. & 1. & 1. & 0.95 & 0.89  \\
\hline
ECGFiveDays & 0.125 & 0.31 & 0.44 & 0.19 & \textbf{0.06} & 0.31 & 0.125 & 1 & 0.125 & 0.84 & \textbf{0.06} \\
\hline
ECG200 & 0.59 & 0.55 & 0.64 & 0.43 & 0.40 & 0.55 & 0.65 & 0.68 & 0.63 & 0.71 & \textbf{0.28}  \\
\hline
Handoutlines & \textbf{0.44} & 0.62 & 0.61 & 0.68 & 0.65 & 0.71 & 0.80 & 0.49 & 0.56 & 0.60 & 0.60 \\
\hline
SonyRobotAI1 & \textbf{0.05} & 0.96 & 0.15 & 0.30 & 0.35 & 0.35 & 1. & 0.25 & 0.25 & 0.80 & 0.40  \\
\hline
SonyRobotAI2 & \textbf{0.25} & 0.90 & 0.85 & 0.45 & 1. & \textbf{0.25} & 1. & 0.9 & 1. & 0.1. & 0.95  \\
\hline
StarLightCurves & \textbf{0.70} & 1. & 1. & 1. & 1. & 1. & 0.92 & 1 & 1 & 1 & 0.8  \\
\hline
TwoLeadECG & 0.43 & 0.43 & 1. & 0.43 & 1. & 0.43 & \textbf{0} & 1. & 1. & 1. & \textbf{0}  \\
\hline
ECG5000 & 0.35 & 0.39 & 0.7 & \textbf{0.08} & 1. & 0.85 & 1. & 0.49 & 0.28 & 0.12 & 0.42 \\
\hline
\end{tabular}}
\caption{FPR at 95\% TPR of different anomaly detection methods calculated on the test set. Bold numbers correspond to the best result (lower is better).}
\label{tab:fpr95tpr}
\end{table}}}

{\renewcommand{\arraystretch}{1.5} 
{\setlength{\tabcolsep}{0.18cm}
\begin{table*}[!htt]
\centering
{\scriptsize
\begin{tabular}{|l|c|c|c|c|c|c|c|c|c|c|c|}
\hline
& SIF & KSIF$_{GW}$ & FIF$_{GW}$ & KSIF$_{C}$ & FIF$_{C}$ & KSIF$_{B}$ & FIF$_{B}$ & IF & OCSVM & fHD & fSDO  \\
\hline
Chinatown & \textbf{1} & 0.99 & 0.92 & 0.99 & 0.98 & \textbf{1} & 0.93 & 0.92 & 0.87 & 0.89 & \textbf{1} \\
\hline
Coffee & 0.90 & \textbf{0.99} & 0.94 & 0.98 & 0.99 & 0.95 & 0.86 & 0.89 & 0.89 & 0.97 & 0.99  \\
\hline
ECGFiveDays & \textbf{0.99} & 0.98 & 0.96 & 0.98 & \textbf{0.99} & \textbf{0.99} & 0.98 & 0.90 & 0.97 & 0.89 & \textbf{0.99}  \\
\hline
ECG200 & \textbf{0.94} & 0.87 & 0.89 & 0.91 & 0.93 & 0.90 & 0.91 & 0.92 & 0.93 & 0.90 & \textbf{0.94}  \\
\hline
Handoutlines & \textbf{0.93} & 0.90 & 0.91 & 0.87 & 0.88 & 0.89 & 0.84 & 0.92 & 0.91 & 0.87 & 0.91  \\
\hline
SonyRobotAI1 & \textbf{0.99} & \textbf{0.99} & \textbf{0.99} & 0.98 & 0.97 & 0.98 & 0.84 & \textbf{0.99} & \textbf{0.99} & 0.88 & \textbf{0.99}  \\
\hline
SonyRobotAI2 & \textbf{0.99} & 0.97 & 0.98 & \textbf{0.99} & 0.91 & \textbf{0.99} & 0.95 & 0.96 & 0.93 & 0.94 & 0.93 \\
\hline
StarLightCurves & \textbf{0.91} & 0.85 & 0.83 & 0.80 & 0.77 & 0.89 & 0.86 & 0.84 & 0.87 & 0.89 & 0.90  \\
\hline
TwoLeadECG & 0.99 & 0.99 & 0.86 & 0.99 & 0.85 & 0.99 & \textbf{1} & 0.88 & 0.74 & 0.82 & \textbf{1}  \\
\hline
ECG5000 & 0.97 & 0.98 & 0.96 & \textbf{0.99} & 0.94 & 0.97 & 0.93 & \textbf{0.99} & \textbf{0.99} & \textbf{0.99} & 0.98 \\
\hline
\end{tabular}}
\caption{AUPR of different anomaly detection methods calculated on the test set. Bold numbers correspond to the best result (higher is better).}
\label{tab:aupr}
\end{table*}}}

{\renewcommand{\arraystretch}{1.5} 
{\setlength{\tabcolsep}{0.18cm}
\begin{table*}[!htt]
\centering
{\scriptsize
\begin{tabular}{|l|c|c|c|c|c|c|c|c|c|c|c|}
\hline
& SIF & KSIF & FIF & IF & OCSVM & fHD & fSDO & DeepSVDD & AnoGan & AutoEncoder & VAE \\
\hline
Chinatown & 0.3 & 0.5 & 1.3 & 0.3 & 0.3 & 0.1 & 0.1 & 18 & 78 & 91 & 78  \\
\hline
Coffee & 0.8 & 1 & 1.6 & 0.3 & 0.3 & 0.1 & 0.1 & 16 & 88 &91 & 90 \\
\hline
ECGFiveDays & 0.6 & 0.6 & 1.2 & 0.5 & 0.6 & 0.3 & 0.3 & 21 & 84 & 87 & 89 \\
\hline
ECG200 & 3 & 5 & 9 & 1 & 1 & 0.7 & 0.7 & 18 & 151 & 91 & 90 \\
\hline
Handoutlines & 30 & 50 & 90 & 7 & 7 & 2 & 2 & 27 &681 & 593 & 365 \\
\hline
SonyRobotAI1 & 0.8 & 0.8 & 1.6 & 0.5 & 0.6 & 0.3 & 0.3 & 18 &84 & 95 & 84  \\
\hline
SonyRobotAI2 & 0.9 & 0.8 & 1.6 & 0.5 & 0.6 & 0.3 & 0.3 & 19 & 86& 92 &82   \\
\hline
StarLightCurves & 13 & 20 & 62 & 3 & 5 & 1 & 1 & 22 & 500& 202& 175 \\
\hline
TwoLeadECG & 0.6 & 0.6 & 1 & 0.3 & 0.3 & 0.2 & 0.2 & 9 & 75 &66 & 58  \\
\hline
ECG5000 & 5.4 & 4 & 15 & 1 & 1 & 0.6 & 0.6  & 9 & 270 & 102 & 91\\
\hline
\end{tabular}}
\caption{Computational time in seconds of different anomaly detection methods calculated on the test set.}
\label{tab:computational_time}
\end{table*}}}

{\renewcommand{\arraystretch}{1.5} 
{\setlength{\tabcolsep}{0.18cm}
\begin{table}[!htt]
\centering
{\scriptsize
\begin{tabular}{|l|c|c|c|c|}
\hline
& DeepSVDD & AnoGan & AutoEncoder & VAE \\
\hline
Chinatown & 0.5 & 0.72 & 1 & 0.99 \\
\hline
Coffee & 0.55 & 0.5 & 0.83 & 0.83 \\
\hline
ECGFiveDays & 0.67 & 0.79 & 0.96 & 0.96 \\
\hline
ECG200 & 0.5 & 0.72 & 0.87 &  0.86 \\
\hline
Handoutlines & 0.63 & 0.55 & 0.82 & 0.82 \\
\hline
SonyRobotAI1 & 0.52 & 0.91 &0.95 & 0.95  \\
\hline
SonyRobotAI2 & 0.64& 0.87& 0.88& 0.88 \\
\hline
StarLightCurves & 0.60 & 0.30 & 0.76 & 0.77 \\
\hline
TwoLeadECG & 0.46 & 0.68 & 1 & 1  \\
\hline
ECG5000 & 0.59 & 0.65 & 0.92 & 0.92\\
\hline
\end{tabular}}
\caption{AUROC of different deep learning based anomaly detection method calculated on the test set (higher is better).}
\label{tab:deep:auc}
\end{table}}}

{\renewcommand{\arraystretch}{1.5} 
{\setlength{\tabcolsep}{0.18cm}
\begin{table}[!htt]
\centering
{\scriptsize
\begin{tabular}{|l|c|c|c|c|}
\hline
& DeepSVDD & AnoGan & AutoEncoder & VAE \\

\hline
Chinatown & 0.76 & 0.89 & 1 & 0.99 \\
\hline
Coffee & 0.80 & 0.77 & 0.94 & 0.94 \\
\hline
ECGFiveDays & 0.93 & 0.96& 0.995 & 0.995 \\
\hline
ECG200 & 0.70 & 0.81 & 0.94 & 0.94 \\
\hline
Handoutlines &  0.73 & 0.63 & 0.86 & 0.86 \\
\hline
SonyRobotAI1 &  0.75& 0.96 & 0.98 &0.98  \\
\hline
SonyRobotAI2 & 0.88 & 0.96& 0.97 & 0.97\\
\hline
StarLightCurves & 0.89& 0.74 & 0.90&0.91 \\
\hline
TwoLeadECG &  0.87& 0.92& 1 & 1 \\
\hline
ECG5000 & 0.93 & 0.91 & 0.99 & 0.99\\
\hline
\end{tabular}}
\caption{AUPR of different deep learning based anomaly detection method calculated on the test set (higher is better).}
\label{tab:deep:aupr}
\end{table}}}

{\renewcommand{\arraystretch}{1.5} 
{\setlength{\tabcolsep}{0.18cm}
\begin{table}[!htt]
\centering
{\scriptsize
\begin{tabular}{|l|c|c|c|c|}
\hline
& DeepSVDD & AnoGan & AutoEncoder & VAE \\

\hline
Chinatown & 0.90 &0.68 & 0 & 0.05 \\
\hline
Coffee & 0.84 & 0.84 &  0.8 & 0.8\\
\hline
ECGFiveDays & 0.70 & 0.65 & 0.5 & 0.5\\
\hline
ECG200 &  0.91 & 0.77 & 0.64 & 0.67\\
\hline
Handoutlines & 0.78 &0.78 & 0.52 & 0.52 \\
\hline
SonyRobotAI1 & 0.88& 0.47 & 0.33 & 0.33  \\
\hline
SonyRobotAI2 & 0.95 & 0.83 & 1 & 1 \\
\hline
StarLightCurves & 0.79&  0.82 & 0.46 & 0.46 \\
\hline
TwoLeadECG & 0.90 & 0.75 & 0 & 0\\
\hline
ECG5000 & 0.85 & 0.63 & 0.68 & 0.68\\
\hline
\end{tabular}}
\caption{FPR at 95\% TPR  of different deep learning based anomaly detection method calculated on the test set (lower is better).}
\label{tab:deep:fpr}
\end{table}}}

\end{document}